\documentclass{article}

\PassOptionsToPackage{numbers, compress}{natbib}

\usepackage[final]{neurips_2022}

\usepackage[utf8]{inputenc} 
\usepackage[T1]{fontenc}    
\usepackage{hyperref}       
\usepackage{url}            
\usepackage{booktabs}       
\usepackage{amsfonts}       
\usepackage{nicefrac}       
\usepackage{microtype}      
\usepackage{xcolor}  
\usepackage{booktabs}       
\usepackage{enumitem}
\usepackage{wrapfig}
\usepackage{multirow}
\usepackage{graphicx}
\usepackage{amsmath}
\usepackage{subcaption}
\usepackage{bbm}
\usepackage{color}
\usepackage{tabularx}
\DeclareMathOperator*{\argmax}{arg\,max}
\usepackage{textcomp}

\newcommand{\etal}{\textit{et al}.}
\definecolor{revise}{RGB}{0,0,225}

\title{Learning to Break the Loop: Analyzing and Mitigating Repetitions for Neural Text Generation 

}

\author{%
  Jin Xu$^1$\thanks{The work was conducted in Apple.},  Xiaojiang Liu$^4$, Jianhao Yan$^2$, Deng Cai$^3$, Huayang Li$^4$, Jian Li$^1$ \\
  $^1$Institute for Interdisciplinary Information Sciences, Tsinghua University \\
  $^2$School of Engineering, Westlake University \\
  $^3$The Chinese University of Hong Kong \\
  $^4$Apple \\
  \texttt{xujin21@mails.tsinghua.edu.cn, xiaojiang\_liu@apple.com, elliottyan37@gmail.com} \\
   \texttt{thisisjcykcd@gmail.com, hli46@apple.com, lijian83@mail.tsinghua.edu.cn} \\
}

\begin{document}

\maketitle

\begin{abstract}
While large-scale neural language models, such as GPT2 and BART,
have achieved impressive results on various text generation tasks, they tend to get stuck in undesirable sentence-level loops with maximization-based decoding algorithms (\textit{e.g.}, greedy search). This phenomenon is counter-intuitive since there are few consecutive sentence-level repetitions in human corpora (e.g., 0.02\% in Wikitext-103). To investigate the underlying reasons for generating consecutive sentence-level repetitions, we study the relationship between the probabilities of the repetitive tokens and
their previous repetitions in the context. Through our quantitative experiments, we find that 1) Language models have a preference to repeat the previous sentence; 2) The sentence-level repetitions have a \textit{self-reinforcement effect}: the more times a sentence is repeated in the context, the higher the probability of continuing to generate that sentence; 3) The sentences with higher initial probabilities usually have a stronger self-reinforcement effect. Motivated by our findings,  we propose a simple and effective training method \textbf{DITTO} (Pseu\underline{D}o-Repet\underline{IT}ion Penaliza\underline{T}i\underline{O}n), where the model learns to penalize probabilities of sentence-level repetitions from pseudo repetitive data. Although our method is motivated by mitigating repetitions, experiments show that DITTO not only mitigates the repetition issue without sacrificing perplexity, but also achieves better generation quality. Extensive experiments on open-ended text generation (Wikitext-103) and text summarization (CNN/DailyMail) demonstrate the generality and effectiveness of our method. Code is released at~\url{https://github.com/Jxu-Thu/DITTO}.
\end{abstract}

\section{Introduction}~\label{sec:intro}
\begin{figure}[t!]
  \hspace*{0.0in}\includegraphics[width=1.0\textwidth]{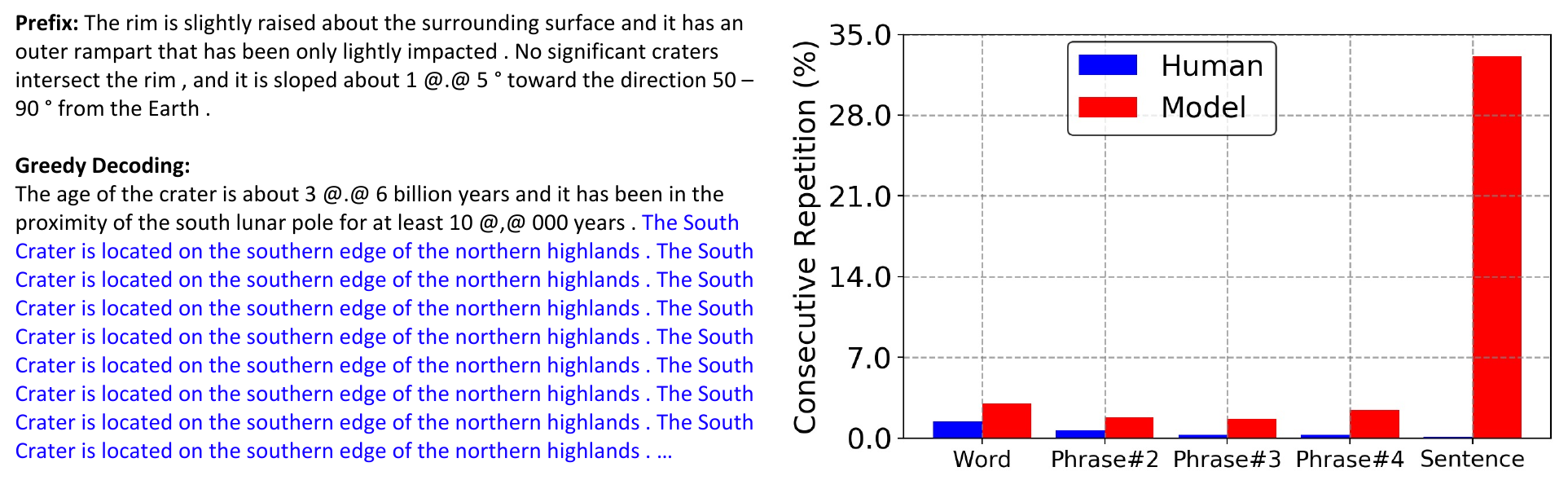} \\
\vspace{-4mm}
\caption{Statistics of human sentences versus model generation on the dev set of Wikitext-103. We train a Transformer model~(750M parameters, similar to GPT-2 Large) on large-scale human corpus Wikitext-103~(over 100 million words). \textbf{Left}: Greedy decoding gets stuck in consecutive sentence-level repetition. \textbf{Right}: The percent of consecutive repetition of the word-, phrase~(\# number of words)- and sentence-level~(see Appendix~\ref{appendix:beam_results} for formulations of consecutive repetition). The model results are the average of generated texts given different prefixes from the Wikitext-103 dev set. Specifically, given 50 tokens as the prefix, the model greedily generates the next 200 tokens. Compared to human language, the model has substantially more consecutive \textit{sentence-level} repetition.}
\label{fig:intro_sen_rep}
\vspace{-5mm}
\end{figure}

Recently, large-scale Transformer-based~\cite{vaswani2017attention} neural language models (e.g., GPT2~\cite{radford2019language}, BART~\cite{lewis2020bart} and OPT~\cite{OPT_fair}) have shown remarkable capability to generate text and achieved better performance than before, such as open-ended generation~\cite{radford2019language,brown2020language} and summarization tasks~\cite{lewis2020bart,OPT_fair}. However, models with standard maximization-based decoding are known to get stuck in redundant consecutive repetitions~\cite{holtzman2019curious}. As shown in Figure~\ref{fig:intro_sen_rep}, the model has a stronger preference for consecutive sentence-level repetitions than the word- or phrase-level while human language has fewer consecutive sentence-level repetitions, which shows a discrepancy between human language and the generated texts. Existing approaches to mitigate repetitions can be categorized into decoding-based and training-based methods. Decoding-based methods rectify these issues by soft or hard $n$-gram blocking~\cite{paulus2018deep, klein2017opennmt}, perplexity/information controlling~\cite{basu2020mirostat, meister2022probability}, and stochastic sampling~\cite{fan2018hierarchical,holtzman2018learning,radford2019language,holtzman2019curious, welleck2020consistency}. Training-based methods minimize the probabilities of tokens that already are generated in the previous context~\cite{welleck2019neural,lin2021straight}. Despite their effectiveness, the reasons why the model prefers repetitions and how repetition occurs during decoding are still unclear.


Fu~\etal~\cite{fu2020theoretical} is the first to analyze the repetition problems from a theoretical perspective by assuming that the language models can be approximated by short-sighted first-order Markov models. However, Holtzman~\etal~\cite{holtzman2019curious} observe the cases of a positive feedback loop of repetitions, which indicates that language models do look at the long-distance context and may not be simply viewed as first-order Markov models. The cases also reveal that the probabilities of repetitive tokens have certain relationships with previous repetitions in context. However, they do not analyze why the model prefers consecutive repetitions. In this paper, we further dig into the problem and conduct \textit{quantitative} experiments to analyze the underlying issue of the repetition.

For a quantitative investigation on consecutive repetitions, we compare the probabilities of the same tokens in repetitive sentences. For example, given a sequence, `I love \textcolor{red}{oranges} . I love \textcolor{blue}{oranges} .', we compare the probability  $\mathcal{P}_{\theta}(\text{`\textcolor{blue}{oranges}'}| \text{`I love \textcolor{red}{oranges} . I love'})$ with $\mathcal{P}_{\theta}(\text{`\textcolor{red}{oranges}'}| \text{`I love'})$. The difference between them is that, for the second `\textcolor{blue}{oranges}', there is already a token `\textcolor{red}{oranges}' that shares the same \textit{sentence-level context} `I love'. We manually repeat the sentence $n$ times as the context so that the next `oranges' has $n$ repetitions in context. In this way, we can investigate the relationship between the probability of the token and the number of repetitions in context. For example, we can first construct the context by repeating the sentence `I love oranges .' $n$ times plus `I love', and then obtain the probability that the model outputs `orange' at the current step.

\begin{figure}[t!]
\vspace{-2mm}
\centering
     \begin{subfigure}[b]{0.48\textwidth}
         \centering
         \includegraphics[width=\textwidth]{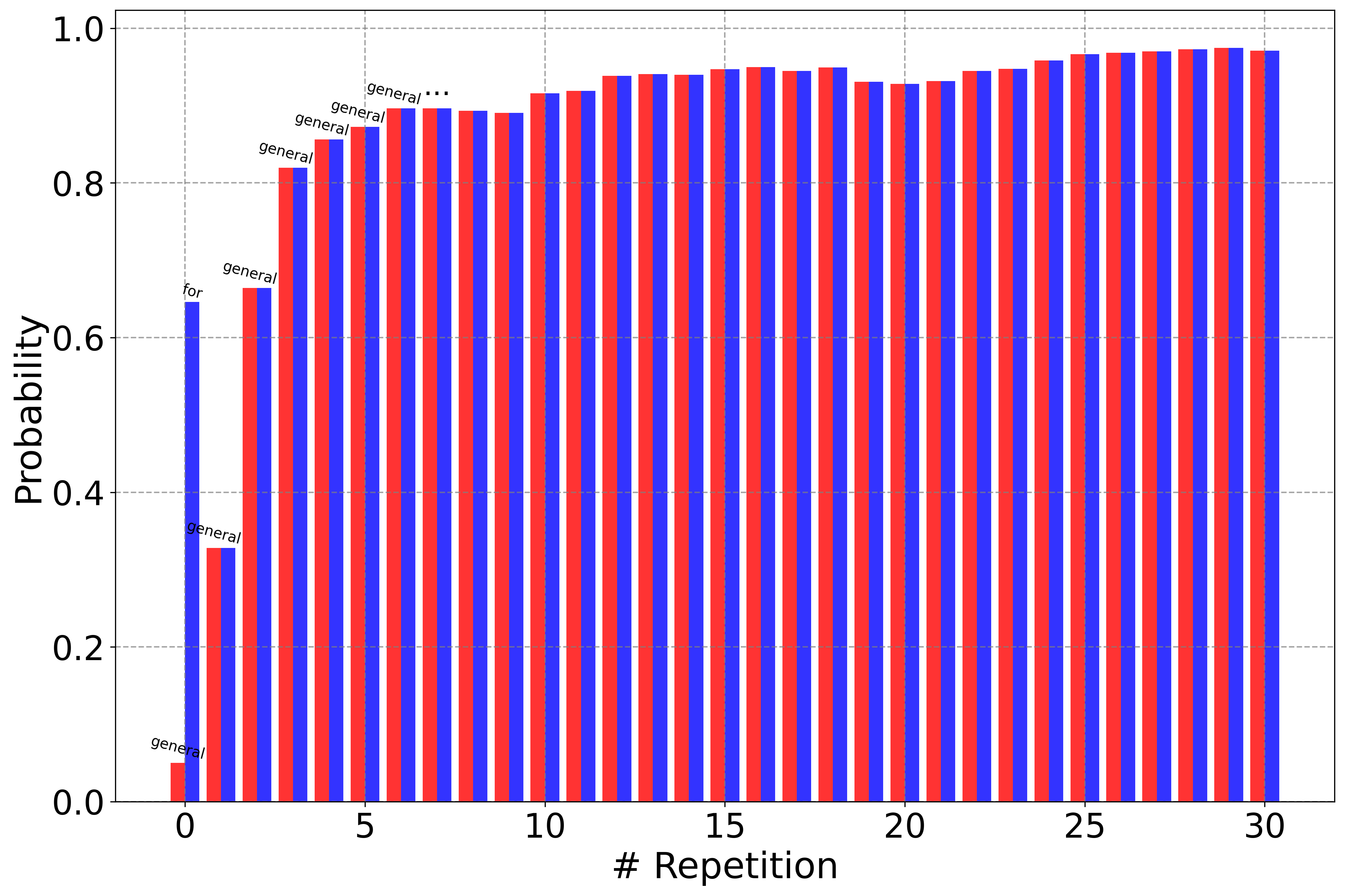}
        \label{fig:analyze_repetition:a}
     \end{subfigure}
     \hfill
     \begin{subfigure}[b]{0.48\textwidth}
         \centering
         \includegraphics[width=\textwidth]{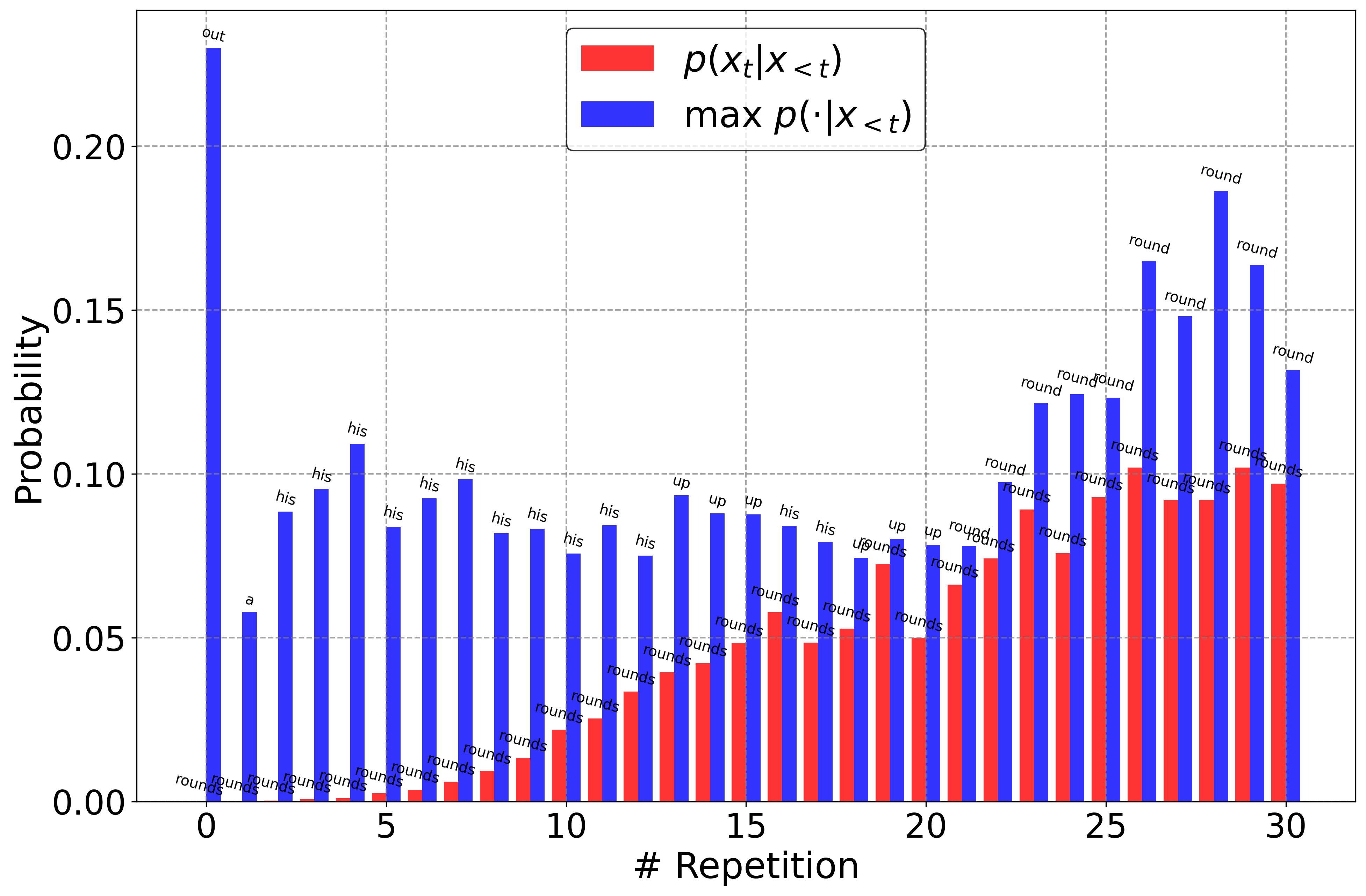}
        \label{fig:analyze_repetition:c}
     \end{subfigure}
        \vspace{-4mm}
        \caption{Manually repeat a given sentence several times, feed to the model and observe the same token's probability $p(x_t|x_{<t})$ (in \textcolor{red}{red}) and maximum probability $\max p(\cdot|x_{<t})$ (in \textcolor{blue}{blue}) as \textit{sentence repetition times increase}. \textbf{Left}: Repeat a normal sentence (`\textit{She is named a proxy \textcolor{red}{general} under Gaara .}') and present the probability of the token `\textcolor{red}{general}'. \textbf{Right}: Repeat a random sentence (`\textit{fría backed \textcolor{red}{rounds} Manganiello Benzedrine Magruder Crego Stansel Zemin compressus .}') where tokens are randomly sampled from the vocabulary, and present the probability of the token `\textcolor{red}{rounds}'. The probability of repetition (in \textcolor{red}{red}) has a self-reinforcement effect: the probability of repetition (\textit{y-axis}) increases almost monotonically with the number of historical repetitions (\textit{x-axis}). Best viewed in color and zoomed in a desktop monitor.}
        \vspace{-4mm}
        \label{fig:self_rein_intro}
\end{figure}

Through our quantitative investigation across the different corpus,
we find that 1) \textbf{The model tends to raise the probability of repeating the previous sentence.} Specifically, even if there is only one sentence-level context repetition, the probability of repetition at the current step increases in most cases. The cause of the phenomenon may be that the model is so confident with its context, when there is a previous token (i.e., `oranges') sharing the same sentence-level context (i.e., `I love'), that the model learns a shortcut to directly copy the token;
2) \textbf{Self-reinforcement effect: the probability of repetition increases almost monotonically with the number of historical repetitions.} Finally, the probability of repetition stabilizes around a certain ceiling value. As shown in Figure~\ref{fig:self_rein_intro}, as the number of repetitions increases, the probability of the word `rounds' and `general' both increase almost monotonically and finally stabilize; 3) \textbf{Sentences with higher initial probabilities usually have a stronger self-reinforcement effect.} For example, in Figure~\ref{fig:self_rein_intro}, we can find that the sentence with a higher initial probability (i.e., the red bar at `0' of x-axis) grows faster and can reach an extremely high value with a few repetitions. Furthermore, the sentences with a higher initial likelihood (e.g., sentences generated by the model itself with maximization-based decoding algorithms) may have a stronger self-reinforcement effect. 

According to our findings, the reasons why the model tends to repeat themselves are as follows: The sentence repetition occurs since the previous sentence generated by the maximization-based decoding algorithms has a relatively high probability, and the model tends to further increase the probability of repeating that sentence. 
Once the model generates one repetitive sentence, the probability of the repetitive sentence would be further enhanced since there are more repetitions sharing the same sentence-level context to support the choice of copying. As a result, the model gets stuck in the sentence-level loops due to the \textit{self-reinforcement effect}.

To mitigate the tendency to repeat previous sentences and overcome the self-reinforcement effect, we aim to make the model aware that the probabilities of generating redundant sentences should be reduced. In addition, the more sentence-level repetitions exist, the lower the probability of the redundant sentence should be. In this paper, we propose a simple and effective training-based method,  Pseu\underline{D}o-Repet\underline{IT}ion Penaliza\underline{T}i\underline{O}n (\textbf{DITTO}). 
We first manually construct pseudo data by repeating a sentence sampled from the training corpus. Then, we design a repetition penalization loss function on these pseudo data for the model to learn to exponentially decay the repetition probability as the number of sentence repetitions increases. The experimental results show that DITTO not only significantly reduces repetitions without sacrificing language model capability in terms of perplexity, but also achieves better generation quality. In the Wikitext-103~\cite{merity2016pointer} open-ended generation task, our methods can achieve lower perplexity and higher MAUVE scores than the strong competitors. With stochastic decoding strategies such as top-$k$~\cite{fan2018hierarchical} and nucleus sampling~\cite{holtzman2019curious}, generated texts by our methods are the closest to human-written texts, as measured by MAUVE score and human evaluation. In the commonly-used CNN/DailyMail summarization benchmark~\cite{hermann2015teaching, nallapati2016abstractive}, our methods also achieve the best performance among the strong competitors.



\section{Analyzing Repetition}\label{sec:analyses}
In this section, we would like to quantitatively investigate the relationship between the probabilities of repetitive sentences and previous repetitions in context. We first define several metrics to measure how the probability changes as the number of repetitions increases. Then, we summarize our findings from the experimental results, and discuss how sentence-level repetitions occur during decoding.

\subsection{Experiment Design}

Formally, we have a sentence $\mathbf{s}$ from a corpus $\mathcal{D}$ and repeat it $N$ times to construct a sample sequence $\mathbf{x}=(\mathbf{s}^0, \mathbf{s}^1, \mathbf{s}^2, \cdots, \mathbf{s}^N)$ where $\mathbf{s}^n=(x_{n,1}, \cdots, x_{n,L_{\mathbf{s}}})$. $x_{n,l}$ is the $l$-th token in the $n$-th repetition of the sentence $\mathbf{s}$ and $L_{{\mathbf{s}}}$ is the number of tokens in sentence $\mathbf{s}$. 
Denote the \textit{sentence-level context} of token $x_{n,l}$ as $\mathbf{x}_{n,<l}=(x_{n,1}, \cdots, x_{n,l-1})$. We define that there is a \textit{sentence-level context repetition} for the token $x_{n,l}$ if and only if there is another same token $x_{i,l}$ in context such that $\mathbf{x}_{i,<l}=\mathbf{x}_{n,<l}$ and $i<n$. For example, for the sequence `I love \textcolor{red}{oranges} . I love \textcolor{blue}{oranges} .' aforementioned, the second token `\textcolor{blue}{oranges}' has a \textit{sentence-level context repetition}. By feeding the sequence $\mathbf{x}$ to a pre-trained model $\mathcal{P}_{\theta}$, we can obtain the probability distribution $\mathcal{P}_{\theta}(x_{n, l}|\mathbf{x}_{<n, l})$ where $\mathbf{x}_{<n,l}=(\mathbf{s}^0,\cdots, \mathbf{s}^{n-1}, \mathbf{x}_{n,<l})$. To study whether $\mathcal{P}_{\theta}(x_{n, l}|\mathbf{x}_{<n, l})$ increases as $n$ increases, we define several metrics as follows:

\begin{itemize} [leftmargin=*]
\item \textbf{Average Token Probability:} $\text{TP}(\mathbf{s^n})=\frac{1}{L_{\mathbf{s}}}\sum_{l=1}^{L_{\mathbf{s}}}{\mathcal{P}_{\theta}(x_{n, l}|\mathbf{x}_{<n,l})}$, which is the average token probability of $n$-th repetitive sentence $\mathbf{s}^n$. $\text{TP}(\mathbf{s^0})$ is the initial probability of tokens in the first sentence.

\item \textbf{Rate of Increased Token Probability}: $\text{IP}(\mathbf{s^n})=\frac{1}{L_{\mathbf{s}}}\sum_{l=1}^{L_{\mathbf{s}}}{\mathbbm{1}(\mathcal{P}_{\theta}(x_{n, l}|\mathbf{x}_{<n,l})>\mathcal{P}_{\theta}(x_{0, l}|\mathbf{x}_{<0,l}))}$ where $\mathbbm{1}$ is the indicator function. We use $\text{IP}(\mathbf{s^n})$ to calculate that, how many probabilities of tokens increase in $\mathbf{s^n}$ compared to the initial ones in $\mathbf{s^0}$.
\item \textbf{Winner Rate:} We say $x_{n,l}$ is a \textit{winner} if  $\mathcal{P}_{\theta}(x_{n, l}|\mathbf{x}_{<n,l}) > \mathcal{P}_{\theta}(x_{0, l}|\mathbf{x}_{<0,l})$ and $x_{n,l}=\argmax \mathcal{P}(\cdot|x_{<n,l})$. Then, we define the winner rate as $\text{WR}(\mathbf{s^n})=\frac{1}{L_{\mathbf{s}}}\sum_{l=1}^{L_{\mathbf{s}}}{\mathbbm{1}(x_{n, l}\text{ is a \textit{winner}})}$. A higher winner rate means that the repetitions are more likely to be generated by a maximization-based decoding algorithm such as greedy decoding.
\end{itemize}

Following the previous work~\cite{welleck2019neural, lin2021straight}, the pre-trained model is a 16-layer Transformer decoder trained on Wikitext-103~\cite{merity2016pointer}. The details are introduced in Sec.~\ref{sec:exp}. Given the corpus $\mathcal{D}$, we can calculate the average values of TP, IP and WR of the $n$-th repetitive sentence $\mathbf{s}^n$ as 
\begin{equation*}
    \text{TP}_n=\frac{1}{|\mathcal{D}|}\sum_{\mathbf{s}\in\mathcal{D}}\text{TP}(\mathbf{s}^n),\quad \text{IP}_n=\frac{1}{|\mathcal{D}|}\sum_{\mathbf{s}\in\mathcal{D}}\text{IP}(\mathbf{s}^n),\quad \text{WR}_n=\frac{1}{|\mathcal{D}|}\sum_{\mathbf{s}\in\mathcal{D}}\text{WR}(\mathbf{s}^n)
\end{equation*}
by enumerating all sentences $\mathbf{s}$ in $\mathcal{D}$. 
To further study the effect of different corpus, we construct three corpus 1) $\mathcal{D}_{\text{random}}$: the tokens of sentences are randomly sampled from the vocabulary of the model; 2) $\mathcal{D}_{\text{book}}$: the sentences are randomly sampled from BookCorpus~\cite{Zhu_2015_ICCV}, and 3) $\mathcal{D}_{\text{wiki}}$: the sentences are randomly sampled from the dev set of Wikitext-103. The size of all different corpus is 1,000.

\begin{figure}[t!]
\vspace{-2mm}
\centering
     \begin{subfigure}[b]{0.325\textwidth}
         \centering
         \includegraphics[width=\textwidth]{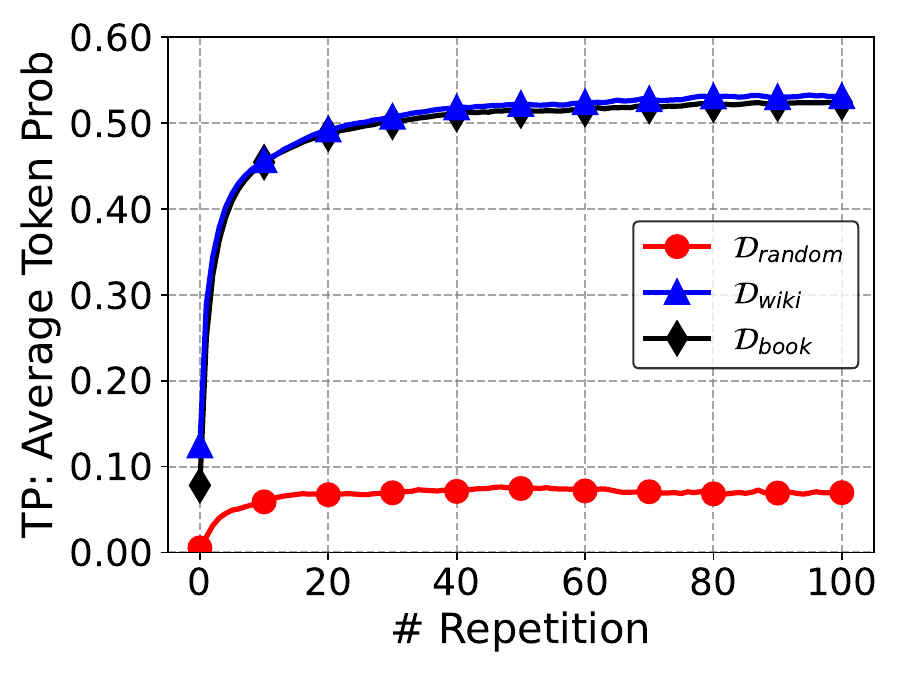}
         \vspace{-7mm}\caption*{(a)}
        \label{fig:analyze_repetition_re:a}
     \end{subfigure}
     \hfill
     \begin{subfigure}[b]{0.325\textwidth}
         \centering
         \includegraphics[width=\textwidth]{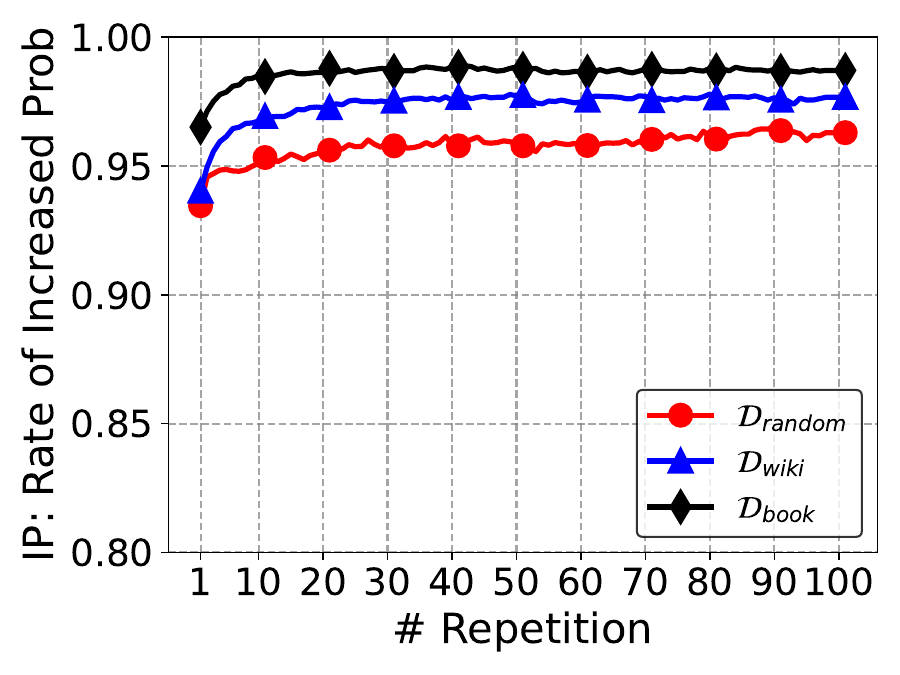}
         \vspace{-7mm} \caption*{(b)}
        \label{fig:analyze_repetition_re:b}
     \end{subfigure}
     \hfill
     \begin{subfigure}[b]{0.325\textwidth}
         \centering
         \includegraphics[width=\textwidth]{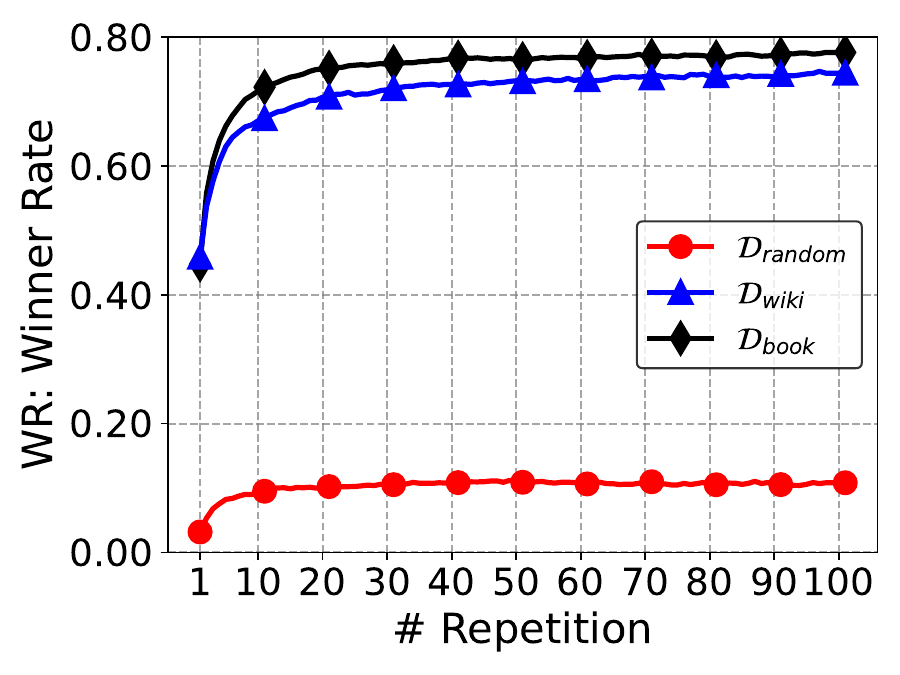}
         \vspace{-7mm}\caption*{(c)}
        \label{fig:analyze_repetition_re:c}
     \end{subfigure}
        \caption{Construct a sample by repeating a sentence $N=100$ times, manually feed it to the model and calculate $\text{TP}_n$ (average token probability), $\text{IP}_n$ (rate of increased probability) and $\text{WR}_n$ (winner rate) for $n=1,\cdots,N$. The results average on sentences from different corpus $\mathcal{D}_{\text{random}}$, $\mathcal{D}_{\text{book}}$ and  $\mathcal{D}_{\text{wiki}}$ respectively. }
        \vspace{-5mm}
        \label{fig:analyze_repetition}
\end{figure}

\subsection{Results and Analyses}
In the analyses, we attempt to answer: 1) Why sentence repetitions occur? 2) Why the model gets stuck into the repetition loop? 3) What kinds of sentences are more likely to be repeated? 

\paragraph{Why sentence repetitions occur?}
As shown in Figure~\ref{fig:analyze_repetition}~(b), IP$_1$ is higher than 90\% across the various corpus, meaning that even if there is only one sentence-level context repetition, the probability of repetition increases in most cases. That indicates \textit{the model has a strong preference to repeat the previous sentence}. Note that the token repetition has not occurred at the current prediction step, and there is only the same sentence-level context. For example, model assigns a higher probability to $\mathcal{P}_{\theta}(\text{`oranges'}|\text{`I love oranges . I love'})$ than $\mathcal{P}_{\theta}(\text{`oranges'}|\text{`I love'})$ since the model has seen the pattern `I love oranges' in the previous context. Thus, the model may be too confident with its previous context repetition and learn a `cheap' shortcut where it directly copies the next token, `oranges'. It is quite different from human language because human would try to avoid full sentence repetition, as the repeated sentence has no new information in most cases.

\paragraph{Why the model gets stuck into the repetition loop?}
Furthermore, as shown in Figure~\ref{fig:analyze_repetition}, as the number of repetitions increases, TP, IP and WR increase monotonically. That means, \textit{the sentence-level repetition has a self-reinforcement effect:} the more times a sentence has been repeated in the context, the higher the probability of continuing to generate that sentence. Finally, IP, WR, TP converge around certain ceiling values. 

\paragraph{What kinds of sentences are more likely to be repeated?}
We can find that 
\textit{the sentences with high initial sentence probability TP$_0$, such as sentences sampled from $\mathcal{D}_{\text{wiki}}$ and $\mathcal{D}_{\text{book}}$,  have a stronger self-reinforcement effect:} TP and WR grows faster and reach the high ceiling values with a few repetitions (See Appendix~\ref{appendix:diff_initial} for more experiments at sentences with different initial probabilities). The higher TP and WR, the more likely the sentence to be repetitively generated by maximization-based decoding algorithms. Note that sentences generated by a maximization-based decoding algorithm have higher initial likelihoods. Thus, the model prefers to repeat itself. 

\paragraph{Analyses}
During decoding with maximization-based algorithms, the sentence-level repetition occurs because 1) Previous generated sentences have high likelihoods and thus have more potential to be repeated; 2) Given previously generated sentences, the model is more likely to generate the repetitive sentence since the model is confident on the historically generated context and tries to find hints in context for the current generation; 3) Once the model repeats the sentence for several times, it would get stuck in the sentence loop due to \textit{self-reinforcement effect}. The effectiveness of stochastic sampling approaches such as top-$k$~\cite{fan2018hierarchical} and nucleus sampling~\cite{holtzman2019curious} may rely on 1) compared to the maximization-based decoding algorithms, the likelihood of previously generated sentences is lower and thus the sentences have less risk of being repeated; 2) current tokens are generated stochastically rather than chosen with maximum probability which can avoid the self-reinforcement effect.

\section{Pseudo-repetition Penalization Training}
According to our analyses in Sec.~\ref{sec:analyses}, it can be clearly seen that the core issue of sentence-level repetition for the model is the tendency to repeat previous sentences and the self-reinforcement effect. In this section, we propose a simple and effective method, named Pseu\underline{D}o-Repet\underline{IT}ion Penaliza\underline{T}i\underline{O}n (\textbf{DITTO}). It first manually feeds repetitive sentences to the model and then explicitly teaches the model to be averse to such repetitions.


\paragraph{Pseudo Repetitive Data Construction.} To construct a pseudo repetitive sample $\mathbf{x}$, we first randomly pick a sentence $\mathbf{s}$ from the training corpus. Then, the pseudo data $\mathbf{x}=(\mathbf{s}^0, \cdots, \mathbf{s}^N)=(x_{0,0},\cdots, x_{1,0}, \cdots, x_{N,0}, \cdots, x_{N,L_s})$ is constructed by repeating the sentence $\mathbf{s}$ by $N+1$ times. The sentences are repeated until they reaches the maximum input sequence length of the model (e.g., 1,536 tokens in the Transformer for open-end generation).

\paragraph{Sentence-level Repetition Penalization.} To define the per-step penalization loss for token $x\in \{x_{1,0}, \cdots, x_{N,L_s}\}$, we define the training objective for the $l$-th token in the $n$-th repetition of the sentence $\mathbf{s}$ as
\begin{equation}\label{eqn:rep_loss}
    \mathcal{L}_{\text{DITTO}}^{n,l}(\mathcal{P}_{\theta}(x_{n,l}|\mathbf{x}_{<n,l}))=-\log(1-\big|\mathcal{P}_{\theta}(x_{n,l}|\mathbf{x}_{<n,l}) - \lambda \cdot \mathcal{P}^*_{\theta}(x_{n-1,l}|\mathbf{x}_{<n-1,l})\big|),
\end{equation}
where $\mathcal{P}_{\theta}^*(\cdot)$ means that the value is excluded for gradient backpropgation, which can implemented by \textit{tensor.detach} in pytorch~\cite{NEURIPS2019_9015}. $\lambda$ is the penalization factor. When $\lambda=1$, the loss function is minimized when the probability of token $x_{n,l}$ in $n$-th repetition is same as that in the $(n-1)$-th repetition to avoid the self-reinforcement effect; when $\lambda<1$, the probability of token $x_{n,l}$ in $n$-th repetition should be smaller than that in the $(n-1)$-th repetition to make model averse to sentence-level repetition. In other words, the probability of tokens in repetitive sentence should decay \textit{exponentially} with a factor of $\lambda$, where $\lambda$ is a hyper-parameter.


In our experiments, we apply the sentence-level repetition penalization by fine-tuning a standard MLE baseline. Fine-tuning is done by equally mixing the sentence-level repetition penalization update and normal MLE loss update. We find that, appending the previous context of the repetitive sentence as the prefix to the pseudo repetitive data $\mathbf{x}$ can achieve better and more stable performance. 
We have tried other alternative loss functions, but they do not perform better. See Appendix~\ref{appendix:loss} for the details.

\paragraph{Discussion}
In human language, there are necessary token-level and phrase-level repetitions that naturally occurs. For example, names of people, city and lemma, set phrases and proverbs appear many times in long documents such as Wikipedia. Thus, given the prefix, the model should have the ability to copy these words or phrases from the prefix and increase their repetition probabilities. These useful repetitions can be viewed as \textit{positive samples}, and our pseudo repetitive data can be viewed as \textit{negative samples}. Combining them for training, the model should learn to distinguish between useful and useless contexts. Thus, although our method is motivated by overcoming sentence-level repetition issues, it may improve the learning ability and 
generalizability of the model. Unlike previous methods (e.g., token-level unlikehood training~\cite{welleck2019neural} and $n$-gram blocking~\cite{paulus2018deep, klein2017opennmt}) that put hard constraints to penalize any repetitions (regardless of the necessity of repetitions), our training objective explicitly encodes the intuition that the model should be inclined to avoid over-repetitions. Therefore, it may alleviate the drawbacks of over-penalization.

\section{Experiments}
\subsection{Setup}\label{sec:exp}
\paragraph{Dataset and Model Training}
We train models on the benchmark dataset Wikitext-103~\cite{merity2016pointer} to evaluate the performance of open-ended generation. The dataset contains over 100 million words. The experiments are conducted at the word level. The model architecture and training hyper-parameters exactly follow the implementations of Welleck~\etal~\cite{welleck2019neural}\footnote{\url{https://github.com/facebookresearch/unlikelihood_training}}. Specifically, we use a 16-layer Transformer with 8 attention heads, hidden size 1024 and fully-connected dimension 4096. There are a total of 750 million parameters, similar to GPT-2 Large~\cite{radford2019language}. We first train the baseline model with standard maximum likelihood (MLE) for a maximum of 150k updates with a batch size of 3 samples. The maximum length of each training sample is 1,536 tokens. Then, we fine-tune the model that has the best validation perplexity with DITTO training for 10k steps. Unless otherwise mentioned, $\lambda$ is set as 0.5. Baseline models are trained following their official implementations. All models are trained on 8 NVIDIA Tesla V100 GPUs. The experiments are implemented based on fairseq codebase~\cite{ott2019fairseq}.

\paragraph{Evaluation}
Following common practices~\cite{welleck2019neural,lin2021straight}, we evaluate the quality of open-ended sequence completions on the test set of Wikitext-103, where the prefix is 50 tokens, and models autoregressively generate the next 100 tokens. Models including various baselines are selected with the best perplexity on the validation set for a fair comparison.
The evaluation metrics are listed as follows:
\begin{itemize}[leftmargin=*]
\item \textbf{MAUVE}: MAUVE~\cite{pillutla2021mauve} is a metric to measure how close model-generated text is to human language by approximately calculating the KL divergence between the distribution of model generated sequences and human language. The range of MAUVE is from 0 to 1. A higher MAUVE score indicates better generated sequences.
\item \textbf{Perplexity and Accuracy}: Given the prefix and true next token, we use perplexity, and next-token prediction accuracy to measure the language modeling ability.
\item \textbf{Repetition}: Following previous work~\cite{welleck2019neural, lin2021straight}, we calculate the portion of duplicate 4-grams (\textbf{Repetition-4}) in a generated sequence to measure phrase-level repetition, defined as $1.0 - |\text{unique 4-grams}|/|\text{4-grams}|$, and average over completions. Similarly, we use portion of duplicate sentences (\textbf{Repetition-Sen}) to measure sentence-level repetition, defined as $1.0 - |\text{unique sentences}|/|\text{sentences}|$. Since there are natural repetitions in human language, the optimal model should produce text whose repetition metrics are \textbf{close to} that of the gold text.
\end{itemize}

\subsection{Results of Open-ended Generation}
\paragraph{Method Comparison} 
We compare with training-based methods since decoding-based methods can be readily applied to models with our training method. Following~\cite{welleck2019neural,lin2021straight}, we first compare different training-based algorithms based on greedy decoding. The results are presented in Table~\ref{table:wiki_deterministic}. Our DITTO can significantly reduce the phrase-level and sentence-level repetitions and achieve the best MAUVE score of 0.77. Although UL-token+seq can generate fewer repetitions, its improvement in the general quality of the model measured by MAUVE is limited since its generations are usually not relevant to the prefix (See Appendix~\ref{appendix:examples} for examples). Compared to the standard baseline MLE, other competitors reduce repetitions by sacrificing perplexity while DITTO achieves lower perplexity and higher accuracy. These results show that DITTO can not only mitigate repetitions but also help the model improve the language modeling and generation quality.

\paragraph{Compatibility with Stochastic Decoding Strategies} Sampling-based decoding algorithms such as top-$k$~\cite{fan2018hierarchical} and top-$p$ (nucleus sampling)~\cite{holtzman2019curious} are widely used in the various applications~\cite{adiwardana2020towards,reynolds2021prompt} and large models~\cite{radford2019language,brown2020language}. We also confirm that our DITTO is compatible with these popular stochastic decoding strategies. The results are presented in Table~\ref{table:wiki_stochastic}. It can be seen that models trained with DITTO can reduce repetitions compared to the MLE baseline and achieve the closest results to human in repetition metrics. For the quality of generated sequences, DITTO achieves the highest MAUVE of 0.96 with top-$k$ or nucleus sampling.
\begin{table}[t]
\centering
\vspace{-6pt}
\caption{Results of different training-based methods on the test set of Wikitext-103 for open-ended generation. 
The results are reported based on three runs with different random seeds. The best value is \textbf{bolded} and the second best is \underline{underlined}.}
\vspace{-0pt}
\label{table:wiki_deterministic}
\scalebox{1.0}{
\begin{tabular}{l|c|cc|cc}
\toprule
\textbf{Model}  & \textbf{MAUVE}   & \textbf{Perplexity} & \textbf{Accuracy} & \textbf{Repetition-4} & \textbf{Repetition-Sen}  \\\midrule
MLE~\cite{radford2019language}   & 0.34$_{\pm \text{0.02}}$ &   \underline{25.68$_{\pm \text{0.04}}$} & 0.39$_{\pm \text{0.00}}$ & 44.20$_{\pm \text{1.43}}$\% & 14.50$_{\pm \text{1.59}}$\% \\
  UL-token~\cite{welleck2019neural}   & 0.57$_{\pm \text{0.01}}$  &  26.98$_{\pm \text{0.12}}$ & 0.39$_{\pm \text{0.00}}$ & 28.30$_{\pm \text{0.78}}$\% & 7.40$_{\pm \text{0.83}}$\% \\
  UL-token+seq~\cite{welleck2019neural}   & 0.48$_{\pm \text{0.03}}$  & 25.95$_{\pm \text{0.08}}$ & \underline{0.40$_{\pm \text{0.00}}$} & \textbf{7.60$_{\pm \text{0.46}}$\%} & \textbf{0.05$_{\pm \text{0.03}}$\%} \\
  SG~\cite{lin2021straight}   & \underline{0.74$_{\pm \text{0.01}}$}  & 25.84$_{\pm \text{0.06}}$ & \underline{0.40$_{\pm \text{0.00}}$} & 23.00$_{\pm \text{0.28}}$\% & 5.24$_{\pm \text{0.75}}$\%  \\
  DITTO (ours)  & \textbf{0.77$_{\pm \text{0.01}}$}  & \textbf{24.33$_{\pm \text{0.04}}$} & \textbf{0.42$_{\pm \text{0.00}}$} & \underline{22.00$_{\pm \text{0.31}}$\%} & \underline{2.85$_{\pm \text{0.74}}$\%}  \\\midrule
  Human & - & - & - & 1.10\% & 0.01\% \\
\bottomrule
\end{tabular}
}
\vspace{-15pt}
\end{table}

\begin{table}[!t]
\centering
\vspace{-1mm}
\caption{Results of different training-based methods on the test set of Wikitext-103 under different stochastic decoding algorithms. $k=50$ and top-$p$ ($p=0.9$) for nucleus sampling. The numbers \textit{closest} to \textit{human scores} are in \textbf{bold} except for MAUVE~\cite{pillutla2021mauve}. }
\label{table:wiki_stochastic}
\scalebox{1.0}{
\begin{tabular}{ll|c|cc}
\toprule
\textbf{Search} & \textbf{Model} & \textbf{MAUVE}  & \textbf{Repetition-4} & \textbf{Repetition-Sen}  \\\midrule
\multirow{5}{*}{Top-$k$} & MLE~\cite{radford2019language} & 0.94$_{\pm \text{0.00}}$  & 1.60$_{\pm \text{0.09}}$\% & 0.25$_{\pm \text{0.06}}$\textperthousand \\
& UL-token~\cite{welleck2019neural} & 0.95$_{\pm \text{0.00}}$  & 0.70$_{\pm \text{0.13}}$\% &  0.00$_{\pm \text{0.00}}$\textperthousand\\
& UL-token+seq~\cite{welleck2019neural} & 0.93$_{\pm \text{0.01}}$  & 0.09$_{\pm \text{0.11}}$\% & 0.06$_{\pm \text{0.02}}$\textperthousand \\
& SG~\cite{lin2021straight} & 0.93$_{\pm \text{0.01}}$  & 0.50$_{\pm \text{0.19}}$\% & 0.00$_{\pm \text{0.00}}$\textperthousand \\
& DITTO & \textbf{0.96$_{\pm \text{0.00}}$}  & \textbf{1.00$_{\pm \text{0.10}}$\%} &  \textbf{0.09$_{\pm \text{0.01}}$\textperthousand}  \\\midrule
\multirow{5}{*}{Nucleus} & MLE~\cite{radford2019language} & 0.94$_{\pm \text{0.00}}$  & 1.40$_{\pm \text{0.08}}$\% & \textbf{0.08$_{\pm \text{0.01}}$\textperthousand}   \\
& UL-token~\cite{welleck2019neural} & 0.94$_{\pm \text{0.00}}$  &  0.47$_{\pm \text{0.08}}$\% &  0.00$_{\pm \text{0.00}}$\textperthousand  \\
& UL-token+seq~\cite{welleck2019neural} & 0.94$_{\pm \text{0.01}}$  & 0.08$_{\pm \text{0.05}}$\% &  0.02$_{\pm \text{0.02}}$\textperthousand  \\
& SG~\cite{lin2021straight} & 0.93$_{\pm \text{0.01}}$ & 0.40$_{\pm \text{0.19}}$\%& 0.06$_{\pm \text{0.01}}$\textperthousand   \\
& DITTO &  \textbf{0.96$_{\pm \text{0.00}}$}  & \textbf{0.98$_{\pm \text{0.09}}$\%} & \textbf{0.08$_{\pm \text{0.01}}$\textperthousand}   \\\midrule
\multicolumn{2}{c|}{Human} & -  & 1.10\% & 0.10\textperthousand \\
\bottomrule

\end{tabular}
}
\vspace{-15pt}
\end{table}

\begin{wraptable}{r}[-2mm]{0.38\textwidth}
\centering
\vspace{-12pt}
\caption{Human Evaluation Results. * means the results are statistically significant (2-sided binomial test, $p$<.05).}
\vspace{-5pt}
\label{table:human}
\scalebox{0.87}{
\begin{tabular}{l|c}
\toprule
 & \textbf{Win Rate}  \\\midrule
\multicolumn{2}{l}{\small\textit{Greedy Search}}\\\midrule
\textbf{DITTO} vs MLE & *84\%\\
\textbf{DITTO} vs UL-token & *80\% \\
\textbf{DITTO} vs UL-token+seq & *65\%\\
\textbf{DITTO} vs SG & *68\% \\\midrule
\multicolumn{2}{l}{\small\textit{Nucleus Sampling $p=0.9$}}\\\midrule
\textbf{DITTO} vs MLE & *65\%\\
\textbf{DITTO} vs UL-token & *71\%\\
\textbf{DITTO} vs UL-token+seq & *62\% \\
\textbf{DITTO} vs SG & *63\%\\
\bottomrule

\end{tabular}
}
\vspace{-18pt}
\end{wraptable}

\paragraph{Human Evaluation} We conduct a pairwise crowdworker evaluation to judge the quality of the generations of DITTO compared to other baselines. For each pair of methods, the model generates texts based on the same 100 random prefixes from the test set of Wikitext-103. Evaluators are asked to independently judge which generation is better in terms of their grammaticality, relevance, coherence and informativeness. The evaluation interface and more details are in Appendix~\ref{appendix:human_eval}. As shown in Table~\ref{table:human}, DITTO consistently and significantly outperforms other competitors across different decoding strategies.

\paragraph{Self-reinforcement Effect}\label{exp:self_rein_study}
As mentioned in Sec.~\ref{sec:analyses}, the sentence-level repetition has a self-reinforcement effect, which is the core issue of sentence-level repetition. We further study it when the model is trained with different methods. The settings of experiments follow those in Sec.~\ref{sec:analyses}. As shown in Figure~\ref{fig:method_repetition_compare}, we can find 1) TP, IP and WR increase almost monotonically in UL and SG, which shows that the self-reinforcement effect has not been solved in these training-based methods;
2) When the model is trained with DITTO, these metrics drop rapidly as the number of repetitions grows, showing the effectiveness of DITTO in overcoming the self-reinforcement effect.


\begin{figure}[t!]
\vspace{-3mm}
\centering
     \begin{subfigure}[b]{0.325\textwidth}
         \centering
         \includegraphics[width=\textwidth]{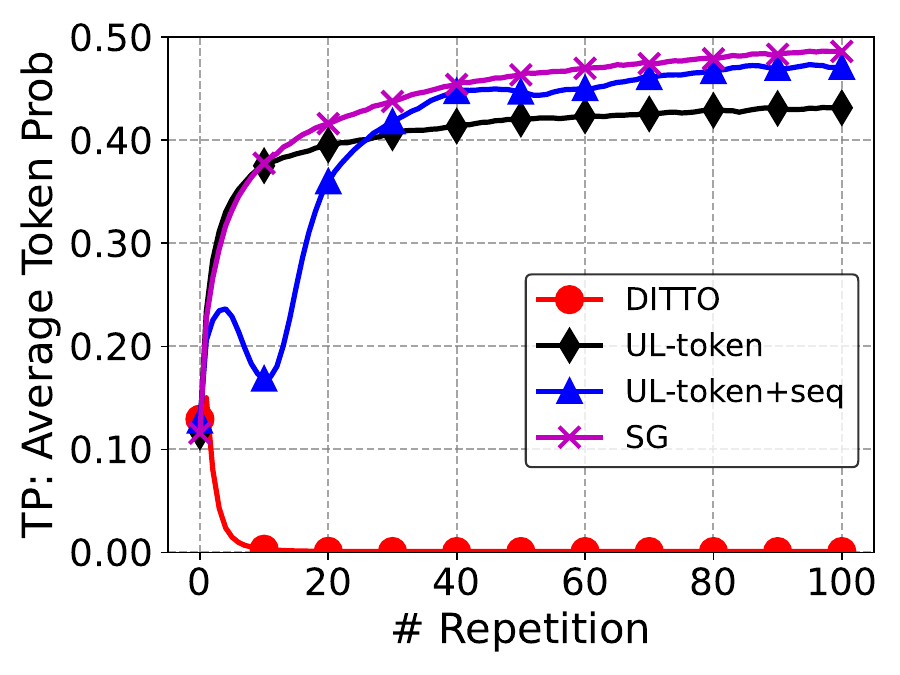}
     \end{subfigure}
     \hfill
     \begin{subfigure}[b]{0.325\textwidth}
         \centering
         \includegraphics[width=\textwidth]{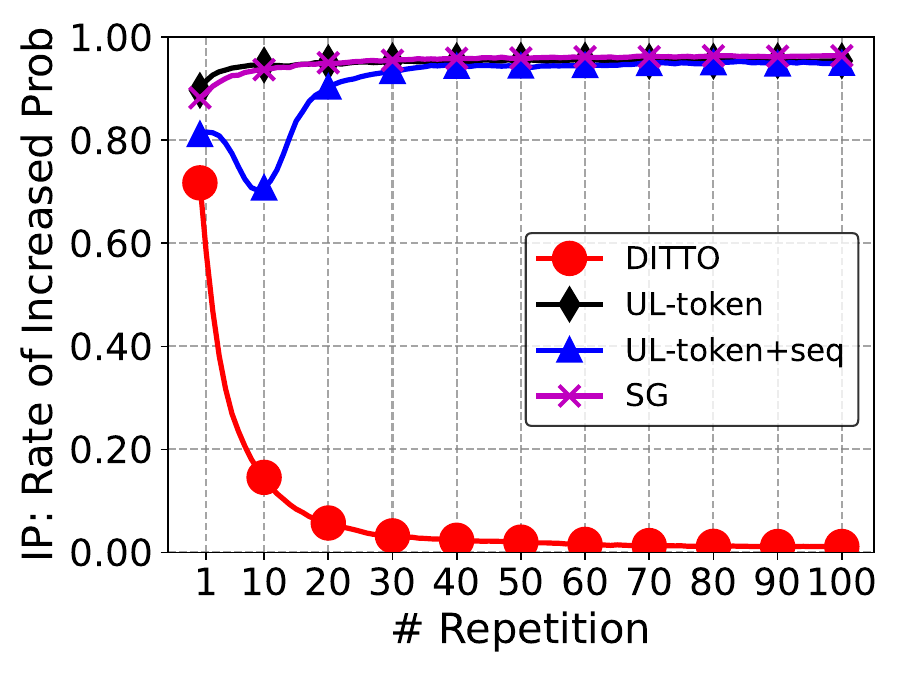}
     \end{subfigure}
     \hfill
     \begin{subfigure}[b]{0.325\textwidth}
         \centering
         \includegraphics[width=\textwidth]{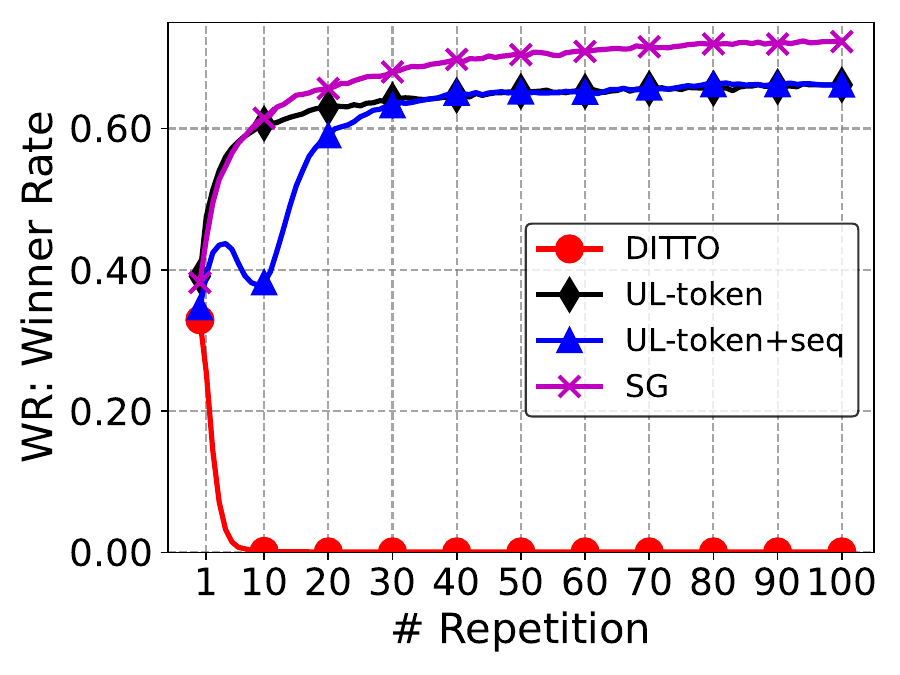}
     \end{subfigure}

        \vspace{-2mm}\caption{Results of different training-based methods by feeding repetitive sentences as described in Sec.~\ref{sec:analyses}]. We average the results on sentences from $\mathcal{D}_{\text{wiki}}$.}
        \label{fig:method_repetition_compare}
        \vspace{-4mm}
\end{figure}

\subsection{Results of Directed Generation}
\begin{table}[ht]
\vspace{-8pt}
\centering
\caption{Abstractive summarization results on CNN/DailyMail.}
\label{table:cnndm}
\scalebox{0.87}{
\begin{tabular}{l|cccc}
\toprule
\textbf{Model} & \textbf{ROUGE-1} & \textbf{ROUGE-2} & \textbf{ROUGE-L} \\
\midrule
Pointer-generator + Coverage~\cite{see2017get} & 39.53 & 17.28 & 36.38 \\
Mask Attention Network~\cite{fan2021mask} & 40.98 & 18.29 & 37.88 \\
BertSum~\cite{liu2019text} & 42.13 & 19.60 & 39.18 \\
UniLM~\cite{dong2019unified} & 43.08 & 20.43 & 40.34 \\
UniLM V2~\cite{bao2020unilmv2} & 43.16 & 20.42 & 40.14 \\
ERNIE-GEN-large~\cite{xiao2021ernie} & 44.02 & 21.17 & 41.26 \\
PEGASUS~\cite{zhang2020pegasus} & 44.17 & 21.47 & 41.11 \\
ProphetNet~\cite{qi2020prophetnet} & 44.20 & 21.17 & 41.30 \\
PALM~\cite{bi2020palm} & 44.30 & 21.12 & 41.14 \\
\midrule
BART-large w.t. MLE~\cite{lewis2020bart} & 44.11$\pm$0.03 & 21.21$\pm$0.01 & 40.83$\pm$0.02\\
BART-large w.t. UL-token~\cite{welleck2019neural} & 44.17$\pm$0.04 & 21.20$\pm$0.02 & 40.83$\pm$0.03 \\
BART-large w.t. UL-token+seq~\cite{welleck2019neural} & 44.13$\pm$0.07 & 21.15$\pm$0.11 & 40.71$\pm$0.09 \\
BART-large w.t. SG~\cite{lin2021straight} & 44.18$\pm$0.06 & 21.17$\pm$0.07 & 40.89$\pm$0.05 \\
\midrule
BART-large w.t. DITTO & \textbf{44.41$\pm$0.03} & \textbf{21.45$\pm$0.01} & \textbf{41.16$\pm$0.02} \\
\bottomrule
\end{tabular}
}
\vspace{-15pt}
\end{table}

\paragraph{Setup}
We further conduct experiments on the directed abstractive summarization task CNN/DailyMail~\cite{hermann2015teaching, nallapati2016abstractive}: given an input document, the model generates several sentences as summarization. We adopt the state-of-the-art model BART-large~\cite{lewis2020bart} as our baseline, which is large-scale encoder-decoder Transformer architecture trained on 160Gb data. 
For DITTO training, given a document and its summarization,  we construct the pseudo data by first random sampling a sentence from the summarization and then repeating it until reaching the maximum summarization length of the decoder model while leaving the document unchanged. We follow the official implementations~\cite{lewis2020bart} 
to train the BART-large model on the CNN/DailyMail and then fine-tune the model with the best validation perplexity using DITTO training. The $\lambda$ is set as 0.9. During inference, tri-gram blocking~\cite{see2017get,nallapati2016abstractive} and beam search (beam size = 5) are used as in Lewis~\etal~\cite{lewis2020bart}.

\paragraph{Results}
We evaluate the performance of the model with the standard F1-based ROUGE~\cite{rouge2004package} scores (ROUGE-1, ROUGE-2, ROUGE-L). As shown in Figure~\ref{table:cnndm}, our training method consistently outperforms UL and SG with a large margin on ROUGE scores and outperforms other competitive baselines. The results also show that DITTO is compatible with the n-gram blocking, a decoding-based method for mitigating repetitions, demonstrating the generality of our approach.

\subsection{Analyses}
\begin{figure}[t!]
  \centering
  \vspace{-3mm}
  \hspace*{-0.05in}\includegraphics[width=1.03\textwidth]{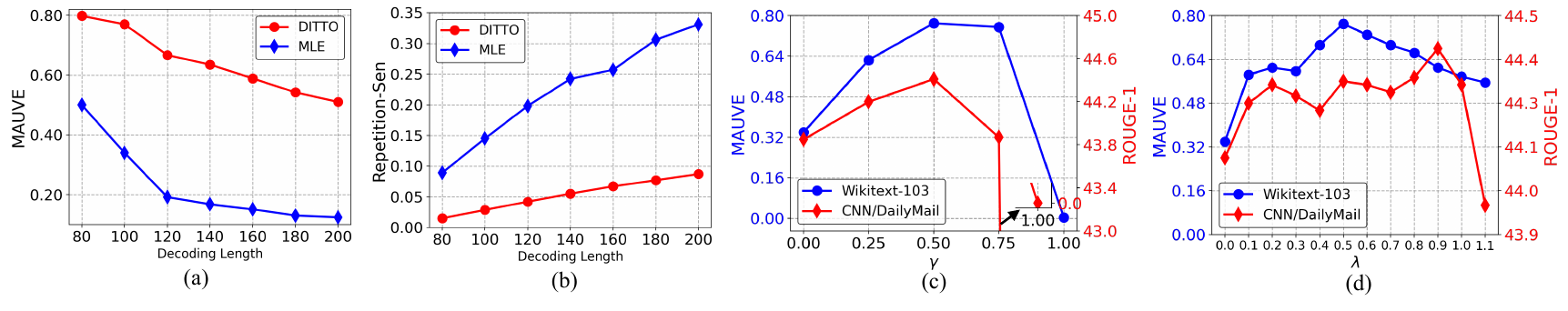} \\
\vspace{-2mm}
\caption{Results of DITTO in different decoding lengths and hyper-parameters. $\lambda$ is the penalization factor and $\gamma$ is the mix ratio of pseudo data and true data. Results are average of three runs with different random seeds.}
\label{fig:para_study}
\vspace{-3mm}
\end{figure}

\paragraph{Auto-completion with different decoding lengths}
From a practice point of view, we analyze DITTO in different decoding lengths. As shown in Figure~\ref{fig:para_study}(a) and (b), the models trained with DITTO consistently and significantly outperform those trained with MLE with the constraints of different decoding lengths, which shows the effectiveness and robustness of our method.

\paragraph{Hyper-parameter Study}
Towards better usage and understanding of DITTO, we study the two hyper-parameters: 1) mix ratio of pseudo data and actual data, denoted as $\gamma$, and 2) penalization factor $\lambda$. As is shown in Figure~\ref{fig:para_study}(c), models achieve the best performance with $\gamma=0.5$ on both datasets. Thus, equally mixing the DITTO update and the MLE training loss update is recommended in practice. As for $\lambda$, we can observe that the optimal value varies in different tasks. In open-ended generation task Wikitext-103, a stronger penalization with $\lambda=0.5$ is preferred. However, a mild penalization with $\lambda=0.9$ is better in the summarization task. 
We conjecture that the repetition issue is more severe in generation tasks with more freedom, such as the open-ended generation.

\section{Related Work}
\paragraph{Repetition in Neural Text Generation.}
Repetition has been a key problem in neural text generation for various tasks including open-ended generation (e.g., text continuation~\cite{holtzman2019curious,welleck2019neural,lin2021straight}) and direct generation (e.g., summarization~\cite{see2017get,liu2019text}). Previous work~\cite{holtzman2019curious,welleck2019neural,li2016simple,karpathy2015deep} observes that, with maximization-based decoding strategies such as greedy search, language models prefer to generate bland and \textit{consecutive} repetitions at the word-level, phrase-level, and sentence-level. Consecutive word-level and phrase-level repetition are lexical repetitions~\cite{guan2021long} where the sentence structure is  incorrect, while sentence-level repetition is semantic repetition that indicates the expression is not informative and has a discrepancy with human language~\cite{guan2021long}. Recently, large-scale pre-training with Transformer architecture~\cite{vaswani2017attention} such as GPT-2~\cite{radford2019language} and BART~\cite{lewis2020bart}, have greatly improved the ability of language modeling and can generate fluent sentences similar to human language. 
However, the generations still have a large number of unexpected consecutive sentence-level repetitions~\cite{radford2019language, brown2020language, fu2020theoretical} in practice, which remains a severe problem in neural text generation.

He~\etal~\cite{he2021exposure} find that, when given the ground-truth context as the prefix or repetitive sentences generated by the model itself as the prefix, the language model can generate high quality texts which shows the model has the self-recovery ability. However, they adopt ancestral sampling to generate continuations rather than maximization-based decoding methods. When the model generates tokens by ancestral sampling such that the current sentence structure is different from previous sentences, the model stops repeating the sentence since their sentence-level context are different. Different from their findings, our analyses reveal that the probability of generation repetitive tokens will increase if they share the same sentence-level context.

\paragraph{Approaches for Mitigating Repetition.} Approaches for mitigating repetition in neural text generation can be categorized into \textit{training-based}~\cite{welleck2019neural,lin2021straight} and \textit{decoding-based}~\cite{see2017get,fan2018hierarchical,holtzman2019curious} approaches. The representative training-based approaches are unlikelihood training (UL)~\cite{welleck2019neural} and straight to gradient (SG)~\cite{lin2021straight}. UL reduces the probability of negative candidate tokens, and SG improves the probability of tokens that do not belong to negative tokens, where the negative tokens are a set of tokens that appear in the previous context. 
However, our experiments in Sec.~\ref{exp:self_rein_study} show that they cannot solve \textit{self-reinforcement} issue. In contrast, our work analyzes why models prefer to repeat themselves and quickly get stuck into the sentence-level loop, and mitigate the self-reinforcement issue. Many decoding-based methods have been proposed to rectify these issues in various tasks. In summarization, $n$-gram blocking~\cite{see2017get,nallapati2016abstractive,lewis2020bart} is often used to block repetitive $n$-grams from subsequent generations. In open-ended tasks, top-$k$~\cite{fan2018hierarchical} and nucleus sampling~\cite{holtzman2019curious, welleck2020consistency} truncate unreliable tail and employ sampling according to token distribution for generating fluent and informative texts. 


\paragraph{Analyses of Repetition Problem.} Although previous work has noticed the repetition issue and many approaches have been proposed to mitigate it at the model training or decoding stages, little has been discussed about the causes of the repetition. Fu~\etal~\cite{fu2020theoretical} theoretically analyzes the problem by assuming the language models can be approximated to first-order Markov models. However, the cases of positive feedback loop of repetitions presented by Holtzman~\etal~\cite{holtzman2019curious} indicate the repetition probability has complex relationships with a quite long-range context and language models may not be simplified as first-order Markov models. However, there is a lack of detailed numeric and statistical results across various sentences and analyses in their work. In contrast, we further dig into the problem and conduct quantitative experiments to analyze underlying issues of repetition.

\section{Conclusion and Future Work}\label{conclusion}
We quantitatively investigate why language models with maximization-based decoding prefer consecutive sentence-level repetitions. Our observations and analyses show that the issue of the repetition is the tendency to repeat previous sentences and the self-reinforcement effect. Guided by our analyses, we propose a simple and effective method named DITTO by constructing pseudo data and teaching model to learn to gradually decay the repetition probability as the number of sentence repetitions grows. The experiments on the Wikitext-103 open-ended generation task and the CNN/DailyMail summarization task demonstrate the superiority of our methods. 

This work investigates the relationship between the probability of the token and the number of repetitions in context. However, there should be deeper reasons for why the model raises the probability of repeating tokens from the perspective of model embedding, neural network architecture or intrinsic characteristics of language. Our current analysis has not touched these aspects, which we leave for future work. We hope our quantitative analyses and approaches to mitigate sentence-level repetitions can help the NLP community better understand, resolve the repetition issue and improve the fundamental language modeling ability of neural language models.

\section{Acknowledgements}
The authors thank three anonymous reviewers and meta reviewer for their comments, which greatly improved the article. Jin Xu and Jian Li are supported in part by the National Natural Science Foundation of China Grant 62161146004, Turing AI Institute of Nanjing and Xi'an Institute for Interdisciplinary Information Core Technology. 

\bibliography{neurips_2022}

\begin{thebibliography}{10}

\bibitem{adiwardana2020towards}
Daniel Adiwardana, Minh-Thang Luong, David~R So, Jamie Hall, Noah Fiedel, Romal
  Thoppilan, Zi~Yang, Apoorv Kulshreshtha, Gaurav Nemade, Yifeng Lu, et~al.
\newblock Towards a human-like open-domain chatbot.
\newblock {\em arXiv preprint arXiv:2001.09977}, 2020.

\bibitem{bao2020unilmv2}
Hangbo Bao, Li~Dong, Furu Wei, Wenhui Wang, Nan Yang, Xiaodong Liu, Yu~Wang,
  Jianfeng Gao, Songhao Piao, Ming Zhou, et~al.
\newblock Unilmv2: Pseudo-masked language models for unified language model
  pre-training.
\newblock In {\em International Conference on Machine Learning}, pages
  642--652. PMLR, 2020.

\bibitem{basu2020mirostat}
Sourya Basu, Govardana~Sachitanandam Ramachandran, Nitish~Shirish Keskar, and
  Lav~R Varshney.
\newblock Mirostat: A neural text decoding algorithm that directly controls
  perplexity.
\newblock In {\em International Conference on Learning Representations}, 2020.

\bibitem{bi2020palm}
Bin Bi, Chenliang Li, Chen Wu, Ming Yan, Wei Wang, Songfang Huang, Fei Huang,
  and Luo Si.
\newblock Palm: Pre-training an autoencoding\&autoregressive language model for
  context-conditioned generation.
\newblock In {\em Proceedings of the 2020 Conference on Empirical Methods in
  Natural Language Processing (EMNLP)}, pages 8681--8691, 2020.

\bibitem{brown2020language}
Tom Brown, Benjamin Mann, Nick Ryder, Melanie Subbiah, Jared~D Kaplan, Prafulla
  Dhariwal, Arvind Neelakantan, Pranav Shyam, Girish Sastry, Amanda Askell,
  et~al.
\newblock Language models are few-shot learners.
\newblock {\em Advances in neural information processing systems},
  33:1877--1901, 2020.

\bibitem{dong2019unified}
Li~Dong, Nan Yang, Wenhui Wang, Furu Wei, Xiaodong Liu, Yu~Wang, Jianfeng Gao,
  Ming Zhou, and Hsiao-Wuen Hon.
\newblock Unified language model pre-training for natural language
  understanding and generation.
\newblock {\em Advances in Neural Information Processing Systems}, 32, 2019.

\bibitem{fan2018hierarchical}
Angela Fan, Mike Lewis, and Yann Dauphin.
\newblock Hierarchical neural story generation.
\newblock In {\em Proceedings of the 56th Annual Meeting of the Association for
  Computational Linguistics (Volume 1: Long Papers)}, pages 889--898, 2018.

\bibitem{fan2021mask}
Zhihao Fan, Yeyun Gong, Dayiheng Liu, Zhongyu Wei, Siyuan Wang, Jian Jiao, Nan
  Duan, Ruofei Zhang, and Xuan-Jing Huang.
\newblock Mask attention networks: Rethinking and strengthen transformer.
\newblock In {\em Proceedings of the 2021 Conference of the North American
  Chapter of the Association for Computational Linguistics: Human Language
  Technologies}, pages 1692--1701, 2021.

\bibitem{fu2020theoretical}
Zihao Fu, Wai Lam, Anthony Man-Cho So, and Bei Shi.
\newblock A theoretical analysis of the repetition problem in text generation.
\newblock {\em arXiv preprint arXiv:2012.14660}, 2020.

\bibitem{guan2021long}
Jian Guan, Xiaoxi Mao, Changjie Fan, Zitao Liu, Wenbiao Ding, and Minlie Huang.
\newblock Long text generation by modeling sentence-level and discourse-level
  coherence.
\newblock In {\em Proceedings of the 59th Annual Meeting of the Association for
  Computational Linguistics and the 11th International Joint Conference on
  Natural Language Processing (Volume 1: Long Papers)}, pages 6379--6393, 2021.

\bibitem{he2021exposure}
Tianxing He, Jingzhao Zhang, Zhiming Zhou, and James Glass.
\newblock Exposure bias versus self-recovery: Are distortions really
  incremental for autoregressive text generation?
\newblock In {\em Proceedings of the 2021 Conference on Empirical Methods in
  Natural Language Processing}, pages 5087--5102, 2021.

\bibitem{hermann2015teaching}
Karl~Moritz Hermann, Tomas Kocisky, Edward Grefenstette, Lasse Espeholt, Will
  Kay, Mustafa Suleyman, and Phil Blunsom.
\newblock Teaching machines to read and comprehend.
\newblock {\em Advances in neural information processing systems}, 28, 2015.

\bibitem{holtzman2019curious}
Ari Holtzman, Jan Buys, Li~Du, Maxwell Forbes, and Yejin Choi.
\newblock The curious case of neural text degeneration.
\newblock In {\em International Conference on Learning Representations}, 2019.

\bibitem{holtzman2018learning}
Ari Holtzman, Jan Buys, Maxwell Forbes, Antoine Bosselut, David Golub, and
  Yejin Choi.
\newblock Learning to write with cooperative discriminators.
\newblock In {\em Proceedings of the 56th Annual Meeting of the Association for
  Computational Linguistics (Volume 1: Long Papers)}, pages 1638--1649, 2018.

\bibitem{karpathy2015deep}
Andrej Karpathy and Li~Fei-Fei.
\newblock Deep visual-semantic alignments for generating image descriptions.
\newblock In {\em Proceedings of the IEEE conference on computer vision and
  pattern recognition}, pages 3128--3137, 2015.

\bibitem{klein2017opennmt}
Guillaume Klein, Yoon Kim, Yuntian Deng, Jean Senellart, and Alexander~M Rush.
\newblock Opennmt: Open-source toolkit for neural machine translation.
\newblock In {\em Proceedings of ACL 2017, System Demonstrations}, pages
  67--72, 2017.

\bibitem{lewis2020bart}
Mike Lewis, Yinhan Liu, Naman Goyal, Marjan Ghazvininejad, Abdelrahman Mohamed,
  Omer Levy, Veselin Stoyanov, and Luke Zettlemoyer.
\newblock Bart: Denoising sequence-to-sequence pre-training for natural
  language generation, translation, and comprehension.
\newblock In {\em Proceedings of the 58th Annual Meeting of the Association for
  Computational Linguistics}, pages 7871--7880, 2020.

\bibitem{li2016simple}
Jiwei Li, Will Monroe, and Dan Jurafsky.
\newblock A simple, fast diverse decoding algorithm for neural generation.
\newblock {\em arXiv preprint arXiv:1611.08562}, 2016.

\bibitem{lin2021straight}
Xiang Lin, Simeng Han, and Shafiq Joty.
\newblock Straight to the gradient: Learning to use novel tokens for neural
  text generation.
\newblock In {\em International Conference on Machine Learning}, pages
  6642--6653. PMLR, 2021.

\bibitem{liu2019text}
Yang Liu and Mirella Lapata.
\newblock Text summarization with pretrained encoders.
\newblock In {\em Proceedings of the 2019 Conference on Empirical Methods in
  Natural Language Processing and the 9th International Joint Conference on
  Natural Language Processing (EMNLP-IJCNLP)}, pages 3730--3740, 2019.

\bibitem{meister2022probability}
Clara Meister, Gian Wiher, Tiago Pimentel, and Ryan Cotterell.
\newblock On the probability-quality paradox in language generation.
\newblock {\em arXiv preprint arXiv:2203.17217}, 2022.

\bibitem{merity2016pointer}
Stephen Merity, Caiming Xiong, James Bradbury, and Richard Socher.
\newblock Pointer sentinel mixture models.
\newblock {\em arXiv preprint arXiv:1609.07843}, 2016.

\bibitem{nallapati2016abstractive}
Ramesh Nallapati, Bowen Zhou, C{\'{\i}}cero~Nogueira dos Santos, {\c{C}}aglar
  G{\"{u}}l{\c{c}}ehre, and Bing Xiang.
\newblock Abstractive text summarization using sequence-to-sequence rnns and
  beyond.
\newblock In Yoav Goldberg and Stefan Riezler, editors, {\em Proceedings of the
  20th {SIGNLL} Conference on Computational Natural Language Learning, CoNLL
  2016, Berlin, Germany, August 11-12, 2016}, pages 280--290. {ACL}, 2016.

\bibitem{ott2019fairseq}
Myle Ott, Sergey Edunov, Alexei Baevski, Angela Fan, Sam Gross, Nathan Ng,
  David Grangier, and Michael Auli.
\newblock fairseq: A fast, extensible toolkit for sequence modeling.
\newblock In {\em Proceedings of NAACL-HLT 2019: Demonstrations}, 2019.

\bibitem{NEURIPS2019_9015}
Adam Paszke, Sam Gross, Francisco Massa, Adam Lerer, James Bradbury, Gregory
  Chanan, Trevor Killeen, Zeming Lin, Natalia Gimelshein, Luca Antiga, Alban
  Desmaison, Andreas Kopf, Edward Yang, Zachary DeVito, Martin Raison, Alykhan
  Tejani, Sasank Chilamkurthy, Benoit Steiner, Lu~Fang, Junjie Bai, and Soumith
  Chintala.
\newblock Pytorch: An imperative style, high-performance deep learning library.
\newblock In H.~Wallach, H.~Larochelle, A.~Beygelzimer, F.~d\textquotesingle
  Alch\'{e}-Buc, E.~Fox, and R.~Garnett, editors, {\em Advances in Neural
  Information Processing Systems 32}, pages 8024--8035. Curran Associates,
  Inc., 2019.

\bibitem{paulus2018deep}
Romain Paulus, Caiming Xiong, and Richard Socher.
\newblock A deep reinforced model for abstractive summarization.
\newblock In {\em International Conference on Learning Representations}, 2018.

\bibitem{pillutla2021mauve}
Krishna Pillutla, Swabha Swayamdipta, Rowan Zellers, John Thickstun, Sean
  Welleck, Yejin Choi, and Zaid Harchaoui.
\newblock Mauve: Measuring the gap between neural text and human text using
  divergence frontiers.
\newblock {\em Advances in Neural Information Processing Systems}, 34, 2021.

\bibitem{qi2020prophetnet}
Weizhen Qi, Yu~Yan, Yeyun Gong, Dayiheng Liu, Nan Duan, Jiusheng Chen, Ruofei
  Zhang, and Ming Zhou.
\newblock Prophetnet: Predicting future n-gram for
  sequence-to-sequencepre-training.
\newblock In {\em Findings of the Association for Computational Linguistics:
  EMNLP 2020}, pages 2401--2410, 2020.

\bibitem{radford2019language}
Alec Radford, Jeff Wu, Rewon Child, David Luan, Dario Amodei, and Ilya
  Sutskever.
\newblock Language models are unsupervised multitask learners.
\newblock 2019.

\bibitem{reynolds2021prompt}
Laria Reynolds and Kyle McDonell.
\newblock Prompt programming for large language models: Beyond the few-shot
  paradigm.
\newblock In {\em Extended Abstracts of the 2021 CHI Conference on Human
  Factors in Computing Systems}, pages 1--7, 2021.

\bibitem{rouge2004package}
Lin~CY ROUGE.
\newblock A package for automatic evaluation of summaries.
\newblock In {\em Proceedings of Workshop on Text Summarization of ACL, Spain},
  2004.

\bibitem{see2017get}
Abigail See, Peter~J Liu, and Christopher~D Manning.
\newblock Get to the point: Summarization with pointer-generator networks.
\newblock In {\em Proceedings of the 55th Annual Meeting of the Association for
  Computational Linguistics (Volume 1: Long Papers)}, pages 1073--1083, 2017.

\bibitem{vaswani2017attention}
Ashish Vaswani, Noam Shazeer, Niki Parmar, Jakob Uszkoreit, Llion Jones,
  Aidan~N Gomez, {\L}ukasz Kaiser, and Illia Polosukhin.
\newblock Attention is all you need.
\newblock {\em Advances in neural information processing systems}, 30, 2017.

\bibitem{welleck2020consistency}
Sean Welleck, Ilia Kulikov, Jaedeok Kim, Richard~Yuanzhe Pang, and Kyunghyun
  Cho.
\newblock Consistency of a recurrent language model with respect to incomplete
  decoding.
\newblock In {\em Proceedings of the 2020 Conference on Empirical Methods in
  Natural Language Processing (EMNLP)}, pages 5553--5568, 2020.

\bibitem{welleck2019neural}
Sean Welleck, Ilia Kulikov, Stephen Roller, Emily Dinan, Kyunghyun Cho, and
  Jason Weston.
\newblock Neural text generation with unlikelihood training.
\newblock In {\em International Conference on Learning Representations}, 2019.

\bibitem{xiao2021ernie}
Dongling Xiao, Han Zhang, Yukun Li, Yu~Sun, Hao Tian, Hua Wu, and Haifeng Wang.
\newblock Ernie-gen: an enhanced multi-flow pre-training and fine-tuning
  framework for natural language generation.
\newblock In {\em Proceedings of the Twenty-Ninth International Conference on
  International Joint Conferences on Artificial Intelligence}, pages
  3997--4003, 2021.

\bibitem{zhang2020pegasus}
Jingqing Zhang, Yao Zhao, Mohammad Saleh, and Peter Liu.
\newblock Pegasus: Pre-training with extracted gap-sentences for abstractive
  summarization.
\newblock In {\em International Conference on Machine Learning}, pages
  11328--11339. PMLR, 2020.

\bibitem{OPT_fair}
Susan Zhang, Stephen Roller, Naman Goyal, Mikel Artetxe, Moya Chen, Shuohui
  Chen, Christopher Dewan, Mona Diab, Xian Li, Xi~Victoria Lin, Todor Mihaylov,
  Myle Ott, Sam Shleifer, Kurt Shuster, Daniel Simig, Punit~Singh Koura, Anjali
  Sridhar, Tianlu Wang, and Luke Zettlemoyer.
\newblock {OPT:} open pre-trained transformer language models.
\newblock {\em CoRR}, abs/2205.01068, 2022.

\bibitem{Zhu_2015_ICCV}
Yukun Zhu, Ryan Kiros, Rich Zemel, Ruslan Salakhutdinov, Raquel Urtasun,
  Antonio Torralba, and Sanja Fidler.
\newblock Aligning books and movies: Towards story-like visual explanations by
  watching movies and reading books.
\newblock In {\em The IEEE International Conference on Computer Vision (ICCV)},
  December 2015.

\end{thebibliography}
\bibliographystyle{plain}

\clearpage
\section*{Checklist}


\begin{enumerate}

\item For all authors...
\begin{enumerate}
  \item Do the main claims made in the abstract and introduction accurately reflect the paper's contributions and scope?
    \answerYes{}
  \item Did you describe the limitations of your work?
    \answerYes{See Section~\ref{conclusion}.}
  \item Did you discuss any potential negative societal impacts of your work?
    \answerNA{}
  \item Have you read the ethics review guidelines and ensured that your paper conforms to them?
    \answerYes{}
\end{enumerate}

\item If you are including theoretical results...
\begin{enumerate}
  \item Did you state the full set of assumptions of all theoretical results?
     \answerNA{}
        \item Did you include complete proofs of all theoretical results?
     \answerNA{}
\end{enumerate}

\item If you ran experiments...
\begin{enumerate}
  \item Did you include the code, data, and instructions needed to reproduce the main experimental results (either in the supplemental material or as a URL)?
   \answerYes{See Section~\ref{sec:exp}. The code is attached in supplemental materials.}
  \item Did you specify all the training details (e.g., data splits, hyperparameters, how they were chosen)?
    \answerYes{See Section~\ref{sec:exp}.}
        \item Did you report error bars (e.g., with respect to the random seed after running experiments multiple times)?
    \answerYes{See reported results in Table~\ref{table:wiki_deterministic},\ref{table:wiki_stochastic} and~\ref{table:cnndm}}
        \item Did you include the total amount of compute and the type of resources used (e.g., type of GPUs, internal cluster, or cloud provider)?
    \answerYes{See Section~\ref{sec:exp}.}
\end{enumerate}

\item If you are using existing assets (e.g., code, data, models) or curating/releasing new assets...
\begin{enumerate}
  \item If your work uses existing assets, did you cite the creators?
    \answerYes{See Section~\ref{sec:exp}.}
  \item Did you mention the license of the assets?
    \answerNA{}
  \item Did you include any new assets either in the supplemental material or as a URL?
    \answerNA{}
  \item Did you discuss whether and how consent was obtained from people whose data you're using/curating?
    \answerNA{}
  \item Did you discuss whether the data you are using/curating contains personally identifiable information or offensive content?
    \answerNA{}
\end{enumerate}

\item If you used crowdsourcing or conducted research with human subjects...
\begin{enumerate}
  \item Did you include the full text of instructions given to participants and screenshots, if applicable?
    \answerYes{See Appendix~\ref{appendix:human_eval}.}
  \item Did you describe any potential participant risks, with links to Institutional Review Board (IRB) approvals, if applicable?
    \answerNA{}
  \item Did you include the estimated hourly wage paid to participants and the total amount spent on participant compensation?
    \answerNA{}
\end{enumerate}

\end{enumerate}

\clearpage
\appendix
\section{Consecutive Repetitions and Statistics of Beam Search Results}~\label{appendix:beam_results}
Previous work~\cite{welleck2019neural, holtzman2019curious} has observed that standard training and greedy decoding usually cause models to generate consecutive repetitive texts. These consecutive repetitive texts are redundant and do not convey new information, which is avoided in human language. There are three types of consecutive repetitions: word-level, phrase-level and sentence-level. The phrase-level means that a phrase consisting of several words is repeated consecutively. The sentence in our paper refers to a sequence split by `.!?' is repeated consecutively~\footnote{The strict definition of the sentence can be found at~\url{https://en.wikipedia.org/wiki/Sentence_(linguistics)}. Here we use the end token to split sentences for ease of experiments.}. We calculate the ratio of consecutive repetition in a sequence $\mathbf{x}$ as follows.

\paragraph{Consecutive word- and phrase-level repetition}
Denote a sequence as $\mathbf{x}=(x_1, \cdots, x_{|\mathbf{x}|})$. The word-level repetition is calculated by $\frac{1}{|\mathbf{x}|-1}\sum_{i=2}^{|\mathbf{x}|}\mathbbm{1}(x_i=x_{i-1})$ where $\mathbbm{1}$ refers to indicator function. The phrase-level repetition where the phrase has $k$ words is calculated by $\frac{1}{|\mathbf{x}|-2k+1}\sum_{i=2k}^{|\mathbf{x}|}\mathbbm{1}((x_{i-k+1}, \cdots, x_{i})=(x_{i-2k+1},\cdots, x_{i-k} ))$. We calculate them for each sequence $\mathbf{x}$ and average over the whole corpus.

\paragraph{Consecutive sentence-level repetition}
Denote a sample sequence as $\mathbf{x}=(\mathbf{s}^0,\cdots,\mathbf{s}^N)$ that contains $(N+1)$ sentences. The sentence-level repetition is calculated by $\frac{1}{N}\sum_{i=1}^{N}\mathbbm{1}(\mathbf{s}^i=\mathbf{s}^{i-1})$.  We calculate it for each sequence $\mathbf{x}$ and average over the whole corpus.

\begin{figure}[htb]
  \centering
  \vspace{-4mm}
  \hspace*{-0.1in}\includegraphics[width=0.5\textwidth]{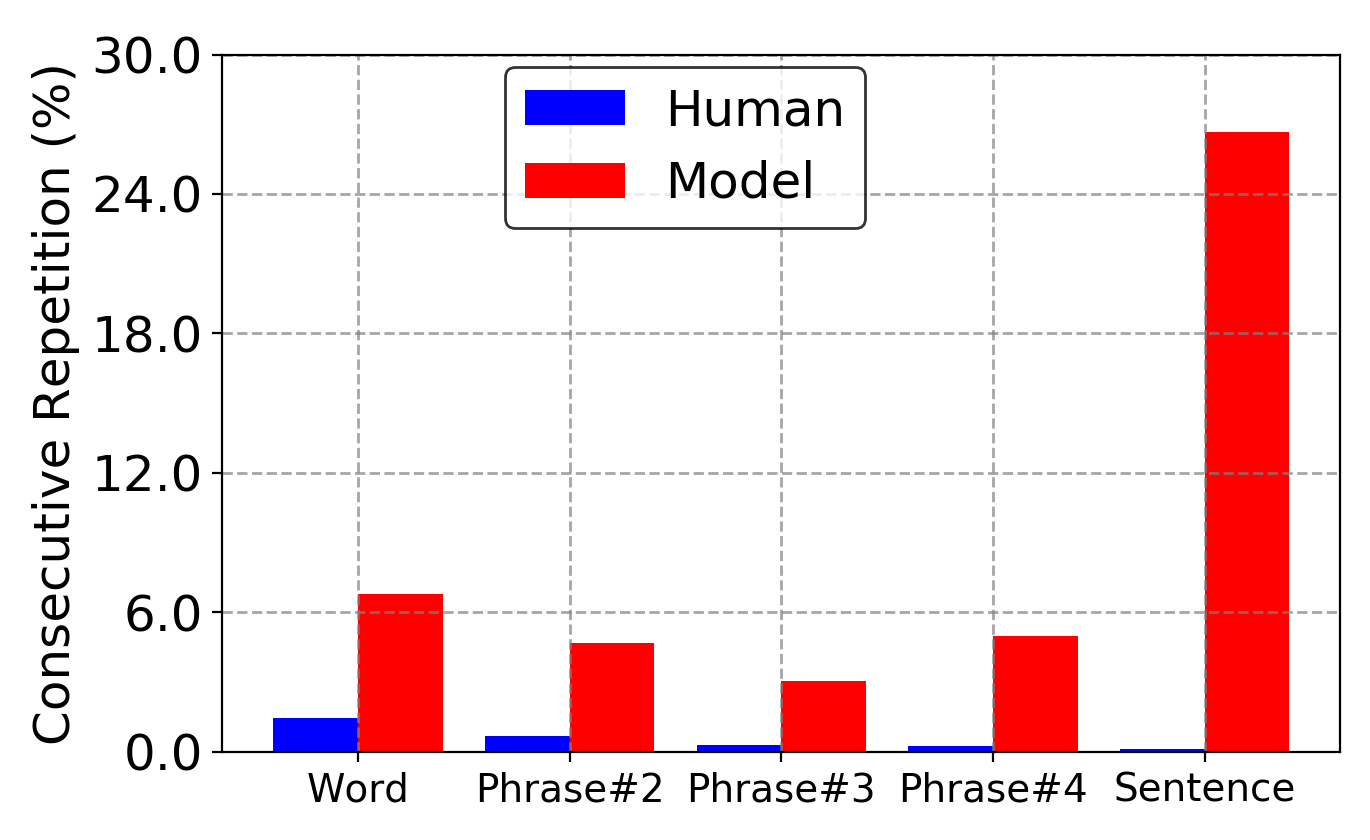} \\
\vspace{-4mm}
\caption{Statistics of human language versus model generation. The model generates the next 200 tokens with beam search ($b=10$) given 50 tokens as the prefix. The results are the average of the Wikitext-103 dev set. Compared to human language, the model has substantially more consecutive \textit{sentence-level} repetition.}
\label{fig:ap_beam_sentence}
\vspace{-5mm}
\end{figure}

As discussed in Section~\ref{sec:intro}, compared to human language, the model with greedy decoding has substantially more sentence-level repetition. This phenomenon holds for other maximization-based decoding methods, such as beam search shown in Figure~\ref{fig:ap_beam_sentence}.

\section{Self-reinforcement in Model Generated Texts}
\begin{figure}[t!]
  \centering
  \hspace*{0.0in}\includegraphics[width=1.0\textwidth]{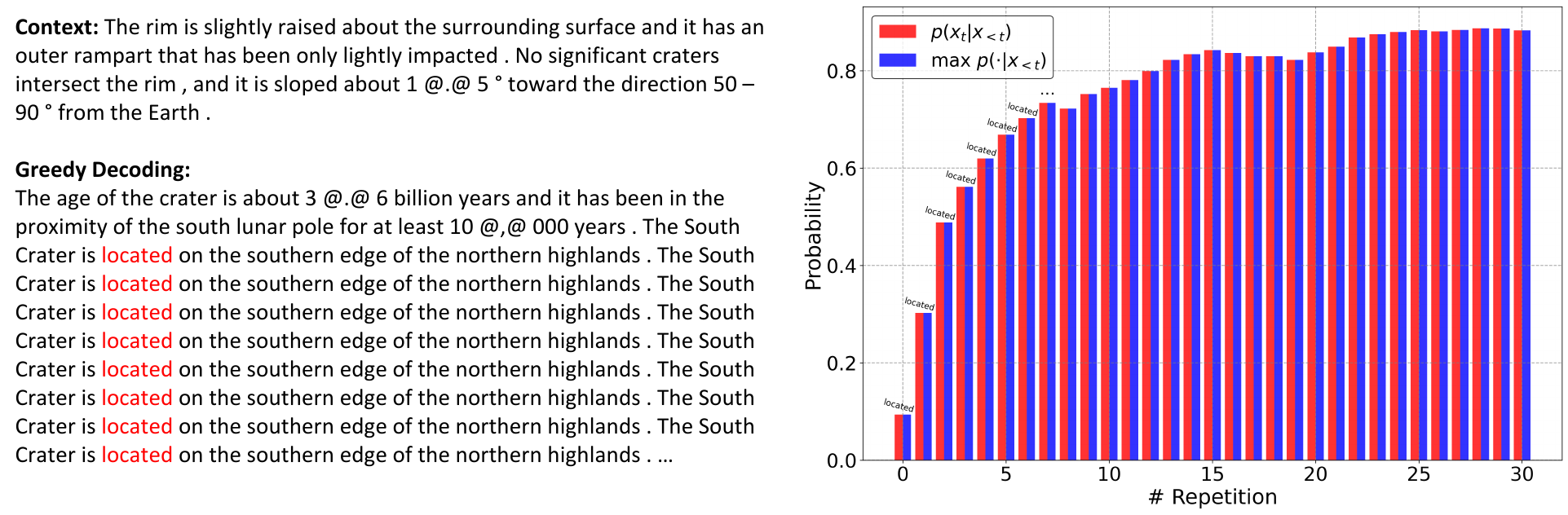} \\
\vspace{-1mm}
\caption{As mentioned in Figure~\ref{fig:intro_sen_rep}, given the prefix, the model gets stuck inthe sentence-level loop (i.e., repeat the sentence `The South Crater is  \textcolor{red}{located} on the southern edge of the northern highlands .'). We present the probability of the token `\textcolor{red}{located}' (\textit{y-axis}) as the number of historical repetitions (\textit{x-axis}) grows. For texts generated by the model autoregressively, the probability of repetition (in \textcolor{red}{red}) also has a \textit{self-reinforcement effect}. Best viewed in color and zoomed in a desktop monitor.}
\label{fig:appendix_generate}
\vspace{-2mm}
\end{figure}

When decoding auto-regressively, the probabilities of the repetitive sentence loops also have a self-reinforcement effect. As shown in Figure~\ref{fig:appendix_generate}, the probability of the token `\textcolor{red}{located}' increases almost monotonically with the number of historical repetitions, which shows the same trend as in Figure~\ref{fig:intro_sen_rep}. 

\paragraph{Why manually repeat sentences to observe the probabilities rather than directly observe those of repetitive texts generated by the model auto-regressively}
Since our target is to study the relationships between the probabilities of repetitive tokens with their repetitions in context, we can manually repeat the sentence and observe the probabilities of the next repetition to simulate the situation when decoding auto-regressively. We do not directly observe the probabilities of repetitions across various generations and calculate metrics TP$_n$, IP$_n$ and WR$_n$, as described in Section~\ref{sec:analyses}, because 1) The model does not always get stuck into the repetitive sentence; 2) There are usually some `dirty' tokens such as `</s>' inserted into the consecutive sentences (e.g., the sentence in Table~\ref{table:example1}) or some minor changes (e.g., the sentence in Table~\ref{table:example2})) in consecutive sentences repetitions so that it is not feasible to calculate those metrics for every token in sentences; 3) Most importantly, we want to study whether the self-reinforcement effect is caused by some problems in the language modeling ability itself so that the effects exists for not only the sentences generated by the model but also any sentences no matter whether the sentences repeated in contexts are grammatically correct.

\section{Loss Function of DITTO}\label{appendix:loss}
The training loss function of DITTO is
\begin{equation}\label{eqn:rep_loss}
    \mathcal{L}_{\text{DITTO}}^{n,l}(\mathcal{P}_{\theta}(x_{n,l}|\mathbf{x}_{<n,l}))=-\log(1-\big|\mathcal{P}_{\theta}(x_{n,l}|\mathbf{x}_{<n,l}) - \lambda \cdot \mathcal{P}^*_{\theta}(x_{n-1,l}|\mathbf{x}_{<n-1,l})\big|),
\end{equation}
where $\mathcal{P}_{\theta}^*(\cdot)$ means that the value is excluded for gradient backpropgation and $\lambda$ is the penalization factor. The loss function is minimized when $\mathcal{P}_{\theta}(x_{n,l}|\mathbf{x}_{<n,l})=\lambda \cdot \mathcal{P}^*_{\theta}(x_{n-1,l}|\mathbf{x}_{<n-1,l})$ where $\lambda$ is a key hyper-parameter. When $\lambda=1$, the loss function requires the probability of token $x_{n,l}$ in $n$-th repetition is same as that in the $(n-1)$-th repetition to avoid the self-reinforcement effect; when $\lambda<1$, the probability of token $x_{n,l}$ in $n$-th repetition should be smaller than that in the $(n-1)$-th repetition to make model averse to sentence-level repetition. To achieve this goal, we can use the following loss function:
\begin{equation}\label{eqn:rep_loss}
    \mathcal{L}_{\text{DITTO-mse}}^{n,l}(\mathcal{P}_{\theta}(x_{n,l}|\mathbf{x}_{<n,l}))=\text{MSE}\big(\mathcal{P}_{\theta}(x_{n,l}|\mathbf{x}_{<n,l}), \lambda \cdot \mathcal{P}^*_{\theta}(x_{n-1,l}|\mathbf{x}_{<n-1,l})\big),
\end{equation}
where MSE refers to mean square error. However, in practice, we find that the MSE loss achieves inferior performance. The results are shown in Table~\ref{table:loss_study}.

\begin{figure}[htb]
  \centering
  \includegraphics[width=0.5\textwidth]{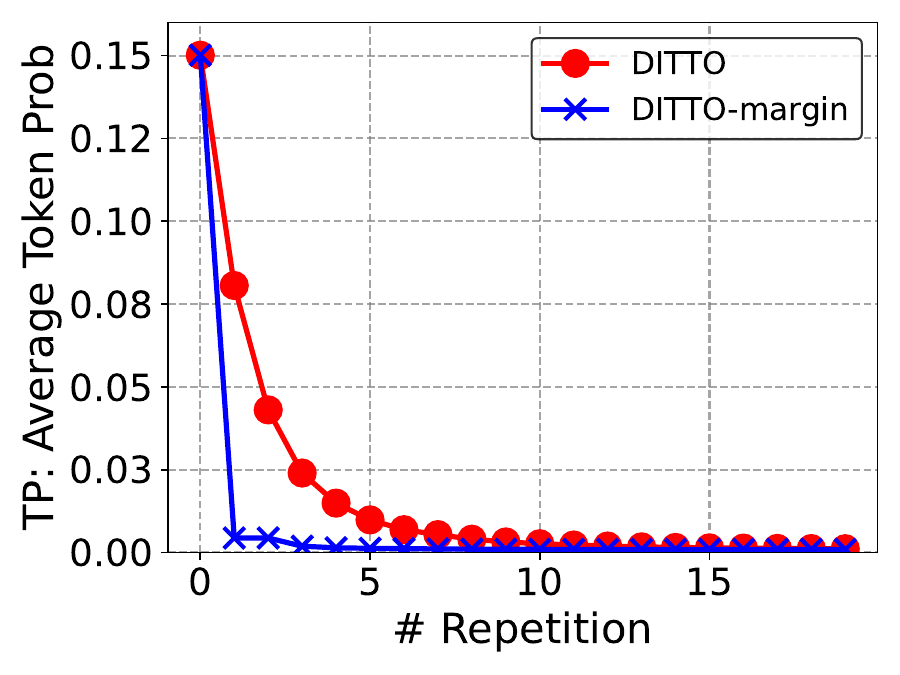} \\
\caption{Comparison of sentence probability. The model trained with $\mathcal{L}_{\text{DITTO-margin}}$ quickly reduces SP to close to 0 even if there is only one sentence repetition.}
\label{fig:margin_analyses}
\end{figure}

Rather than requiring the probability of token $x_{n,l}$ in $n$-th repetition is the same as that in the last repetition multiplying $\lambda$, we can relax the constraint and enable the probability of token $x_{n,l}$ to be no more than that in the $(n-1)$-th repetition multiplying $\lambda$ as

\begin{equation}
    \mathcal{L}_{\text{DITTO-margin}}^{n,l}(\mathcal{P}_{\theta}(x_{n,l}|\mathbf{x}_{<n,l}))= 
\begin{cases}
    0, &\text{if } \mathcal{P}_{\theta}(x_{n,l}|\mathbf{x}_{<n,l}) \leq \lambda \cdot \mathcal{P}^*_{\theta}(x_{n-1,l}|\mathbf{x}_{<n-1,l})\\
    \mathcal{L}_{\text{DITTO}}^{n,l}, &\text{otherwise} 
\end{cases}.
\end{equation}
In practice, as shown in Table~\ref{table:loss_study}, $\mathcal{L}_{\text{DITTO-margin}}$ achieve worse results. We further analyze its performance on the self-reinforcement effect. As shown in Figure~\ref{fig:margin_analyses}, the model trained with $\mathcal{L}_{\text{DITTO-margin}}$ quickly reduces TP to close to 0 even if there is only one sentence repetition. It indicates that the model may have learned a `cheap' solution to optimize the loss function $\mathcal{L}_{\text{DITTO-margin}}$: Regardless of the probability of the previous sentence $\mathcal{P}^*_{\theta}(x_{n-1,l}|\mathbf{x}_{<n-1,l})$, the probability of repeating the previous sentence is directly reduced to 0. Thus, the model may over-penalize repetitions so that all repetitions are forbidden in generations. However, there are some necessary repetitions that naturally occur, as discussed in Section~\ref{sec:analyses}. Over-penalizing repetitions may hurt the language modeling ability and thus lead to inferior performance.

\begin{table}[htb]
\centering
\caption{Results of different loss functions on the test subset of Wikitext-103 for the open-ended generation task.}
\label{table:loss_study}
\scalebox{1.0}{
\begin{tabular}{l|c|cc|cc}
\toprule
\textbf{Objective}  & \textbf{MAUVE}  &  \textbf{Perplexity} & \textbf{Accuracy} & \textbf{Repetition-4} & \textbf{Repetition-Sen} \\\midrule
$\mathcal{L}_{\text{DITTO-mse}}$  & 0.73$_{\pm \text{0.01}}$  & 24.34$_{\pm \text{0.04}}$ & 0.42$_{\pm \text{0.00}}$ & 22.70$_{\pm \text{0.34}}$\% & 2.98$_{\pm \text{0.77}}$\% \\
$\mathcal{L}_{\text{DITTO-margin}}$  & 0.66$_{\pm \text{0.04}}$  & 24.38$_{\pm \text{0.04}}$ & 0.41$_{\pm \text{0.00}}$ & \textbf{19.31$_{\pm \text{0.44}}$\%} & \textbf{2.17$_{\pm \text{0.77}}$\%}  \\
  $\mathcal{L}_{\text{DITTO}}$  & \textbf{0.77$_{\pm \text{0.01}}$}  & \textbf{24.33$_{\pm \text{0.04}}$} & \textbf{0.42$_{\pm \text{0.00}}$} & 22.00$_{\pm \text{0.31}}$\% & 2.85$_{\pm \text{0.74}}$\%  \\\midrule
  Human & -  & - & - & 1.10\% & 0.01\% \\
\bottomrule
\end{tabular}
}
\end{table}


\section{Additional Analyses and Experiments}
\subsection{Self-reinforcement at sentences with different initial probabilities}\label{appendix:diff_initial}

\begin{figure}[t!]
\centering
     \begin{subfigure}[b]{0.48\textwidth}
         \centering
         \includegraphics[width=\textwidth]{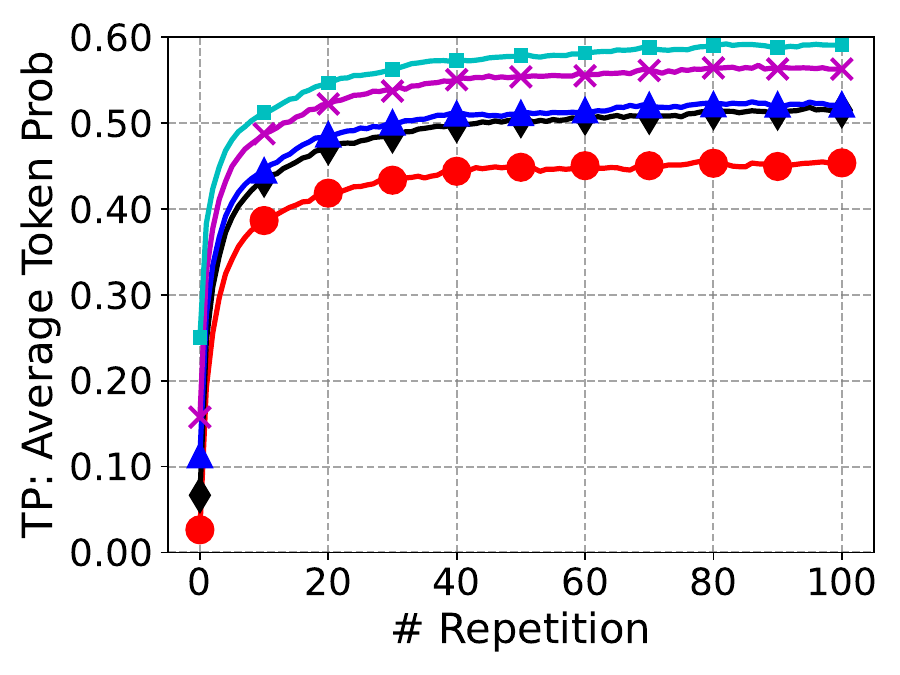}
     \end{subfigure}
     \hfill
     \begin{subfigure}[b]{0.48\textwidth}
         \centering
         \includegraphics[width=\textwidth]{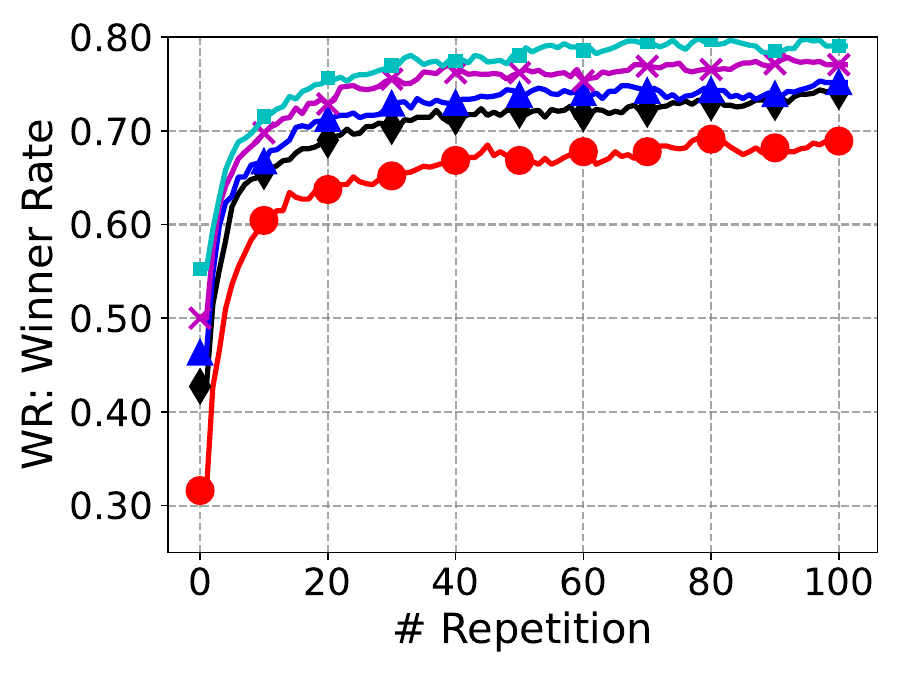}
     \end{subfigure}
     \caption{Results of TP and WR by feeding repetitive sentences as described in Sec.~\ref{sec:analyses}. We equally divide the $\mathcal{D}_{\text{wiki}}$ into 5 groups according to their initial probabilities $\text{TP}(\mathbf{s^0})$ and then average the results.}
        \label{fig:analyses_initial}
\end{figure}

To further verify that sentences with higher initial probabilities usually have a stronger self-reinforcement effect, we equally divide the $\mathcal{D}_{\text{wiki}}$ into 5 groups according to their initial probabilities $\text{TP}(\mathbf{s^0})$ and then repeat the experiments as described in Section~\ref{sec:analyses}. As shown in Figure~\ref{fig:analyses_initial}, sentences with higher initial probabilities reach higher TP and WR as the number of repetitions increases, meaning that these sentences are more likely to be repeated (\textbf{stronger self-reinforcement effect}). For decoding scenarios, if maximization-based decoding algorithms such as greedy decoding are employed, previously generated sentences have a higher initial likelihood (since these sentences themselves are selected with the maximization criterion) and thus, have higher probabilities of being repeated.

\subsection{Study of Pseudo Repetitive Data}
In this section, we study other different choices to construct pseudo repetitive data.

\begin{table}[t]
\centering
\caption{Results of DITTO with the group-level repetitive data. The results are reported based on three runs with different random seeds on the test set of Wikitext-103. }
\label{table:ditto_variant_a}
\scalebox{1.0}{
\begin{tabular}{l|c|cc|cc}
\toprule
\textbf{Model}  & \textbf{MAUVE}   & \textbf{Perplexity} & \textbf{Accuracy} & \textbf{Repetition-4} & \textbf{Repetition-Sen}  \\\midrule
  DITTO w.t. Two Sentences  & 0.77$_{\pm \text{0.01}}$  & 24.37$_{\pm \text{0.03}}$ & 0.42$_{\pm \text{0.00}}$ & 24.20$_{\pm \text{0.38}}$\% & 2.99$_{\pm \text{0.89}}$\%  \\
  DITTO  & 0.77$_{\pm \text{0.01}}$  & 24.33$_{\pm \text{0.04}}$ & 0.42$_{\pm \text{0.00}}$ & 22.00$_{\pm \text{0.31}}$\% & 2.85$_{\pm \text{0.74}}$\%  \\\midrule
  Human & - & - & - & 1.10\% & 0.01\% \\
\bottomrule
\end{tabular}
}
\end{table}

\paragraph{Repeating Two Sentences as Pseudo Data} Rather than repeating one sentence to construct pseudo data, we can randomly pick two consecutive sentences as a group and construct pseudo data by repeating the group. Then, we apply repetition penalization as Eqn.~\ref{eqn:rep_loss} at the group level. As shown in Table~\ref{table:ditto_variant_a}, repeating two sentences as pseudo data for training achieves similar results as the original one.

\begin{figure}[t!]
\centering
     \begin{subfigure}[b]{0.325\textwidth}
         \centering
         \includegraphics[width=\textwidth]{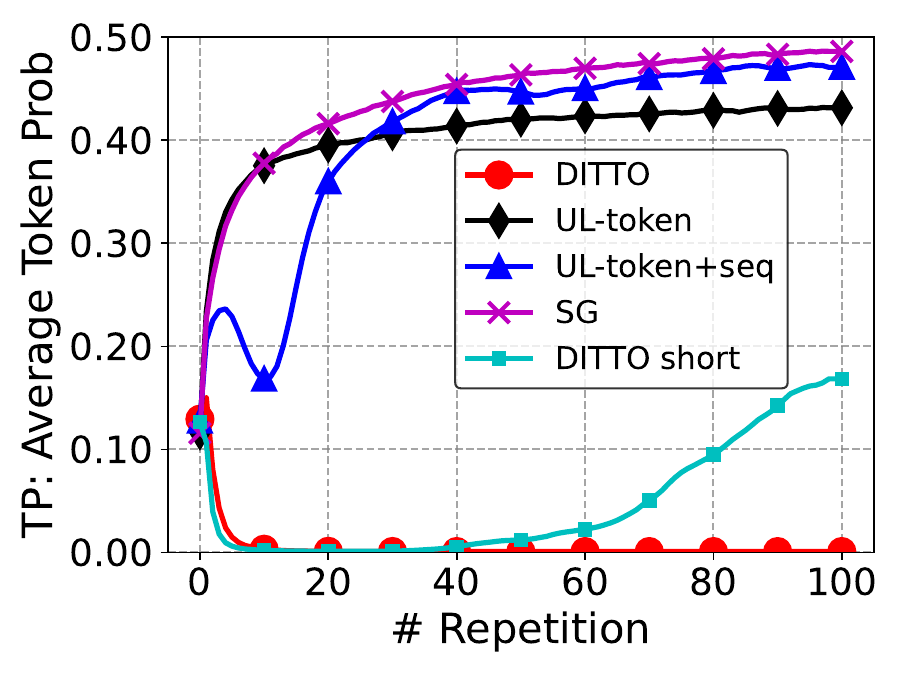}
     \end{subfigure}
     \hfill
     \begin{subfigure}[b]{0.325\textwidth}
         \centering
         \includegraphics[width=\textwidth]{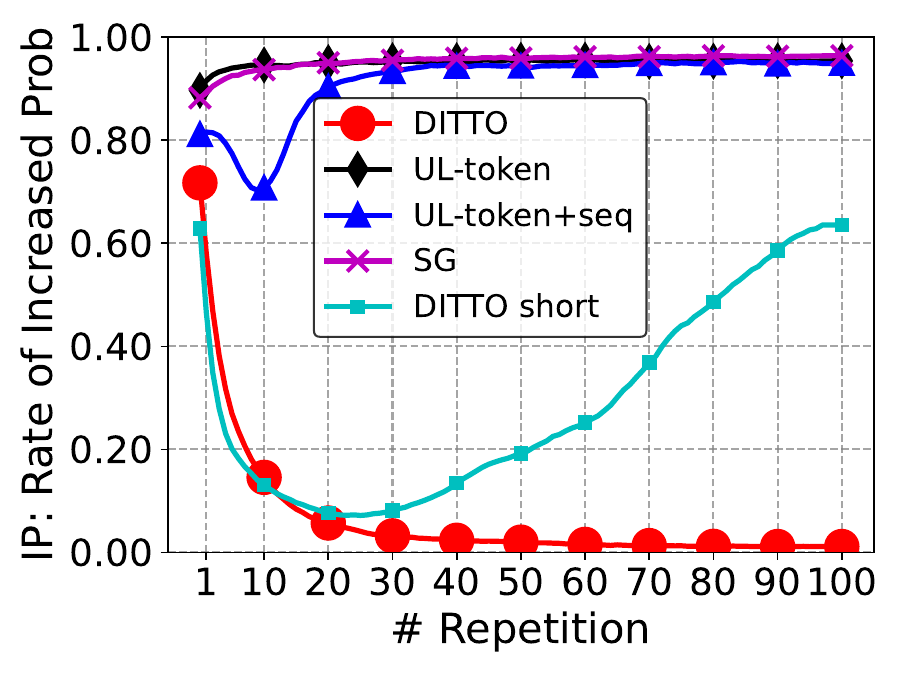}
     \end{subfigure}
     \hfill
     \begin{subfigure}[b]{0.325\textwidth}
         \centering
         \includegraphics[width=\textwidth]{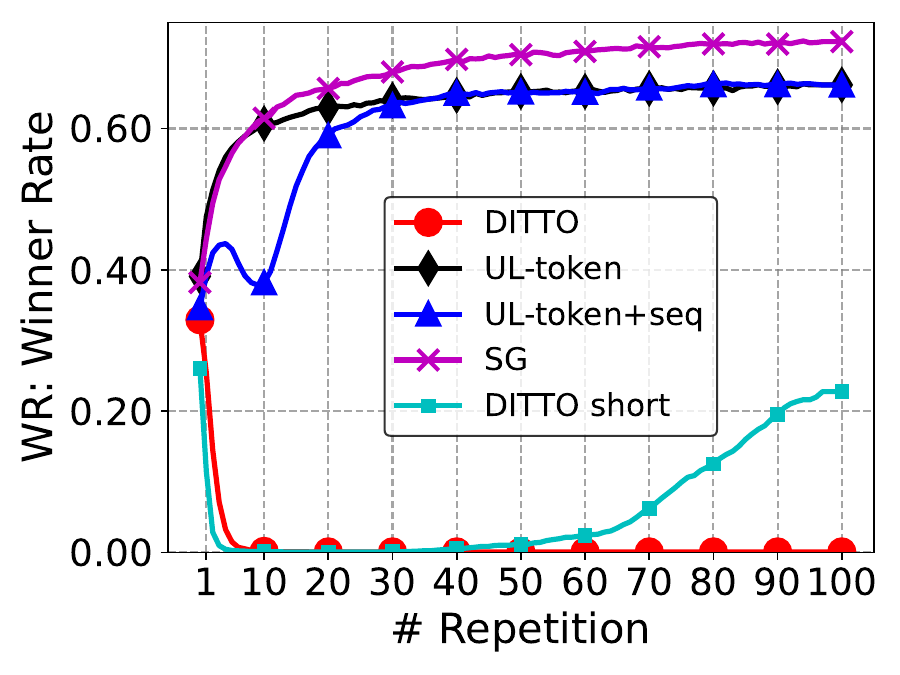}
     \end{subfigure}

        \vspace{-2mm}\caption{Results of different training-based methods by feeding repetitive sentences. `DITTO short' means we only feed repetitive sentences with a maximum input length of 150 tokens for training. We average the results on sentences from $\mathcal{D}_{\text{wiki}}$.}
        \label{fig:method_repetition_compare}
        \vspace{-4mm}
\end{figure}


\paragraph{Length of Pseudo Repetitive Data}
DITTO constructs pseudo repetitive data by repeating sentences until reaching the maximum input sequence length of the model (e.g., 1536 tokens). To further study whether the length of pseudo data has an effect on overcoming the self-reinforcement effect, we short the maximum input length of the repetitive sequence to 150 tokens for training, named DITTO-short. Then, we measure the TP, IP, and WR metrics. The results are shown in Figure~\ref{fig:method_repetition_compare}. From Figure~\ref{fig:method_repetition_compare}, we can see that, for the short-decoding lengths (e.g., generating the next 100 tokens in open-ended generation tasks), DITTO-short can effectively reduce the values of TP, IP and WR. However, for long decoding length, TP, IP, and WR gradually increase at the end,  which does not share the case with DITTO, showing that long sequence penalization is necessary to overcome the self-reinforcement effect. Compared with UL-token+seq~\cite{welleck2019neural}, which also uses 150 tokens for penalization, DITTO enjoys two benefits: 1) DITTO can directly feed longer sequences for penalization training without significantly increasing the computational cost while UL-token+seq needs auto-regressive generation; 2) with the same penalization length (150 tokens), DITTO is more effective on overcoming self-reinforcement effect (e.g., lower TP, IP and WR values).

\section{Human Evaluation Details}\label{appendix:human_eval}
\begin{figure}[htb]
  \centering
  \includegraphics[width=\textwidth]{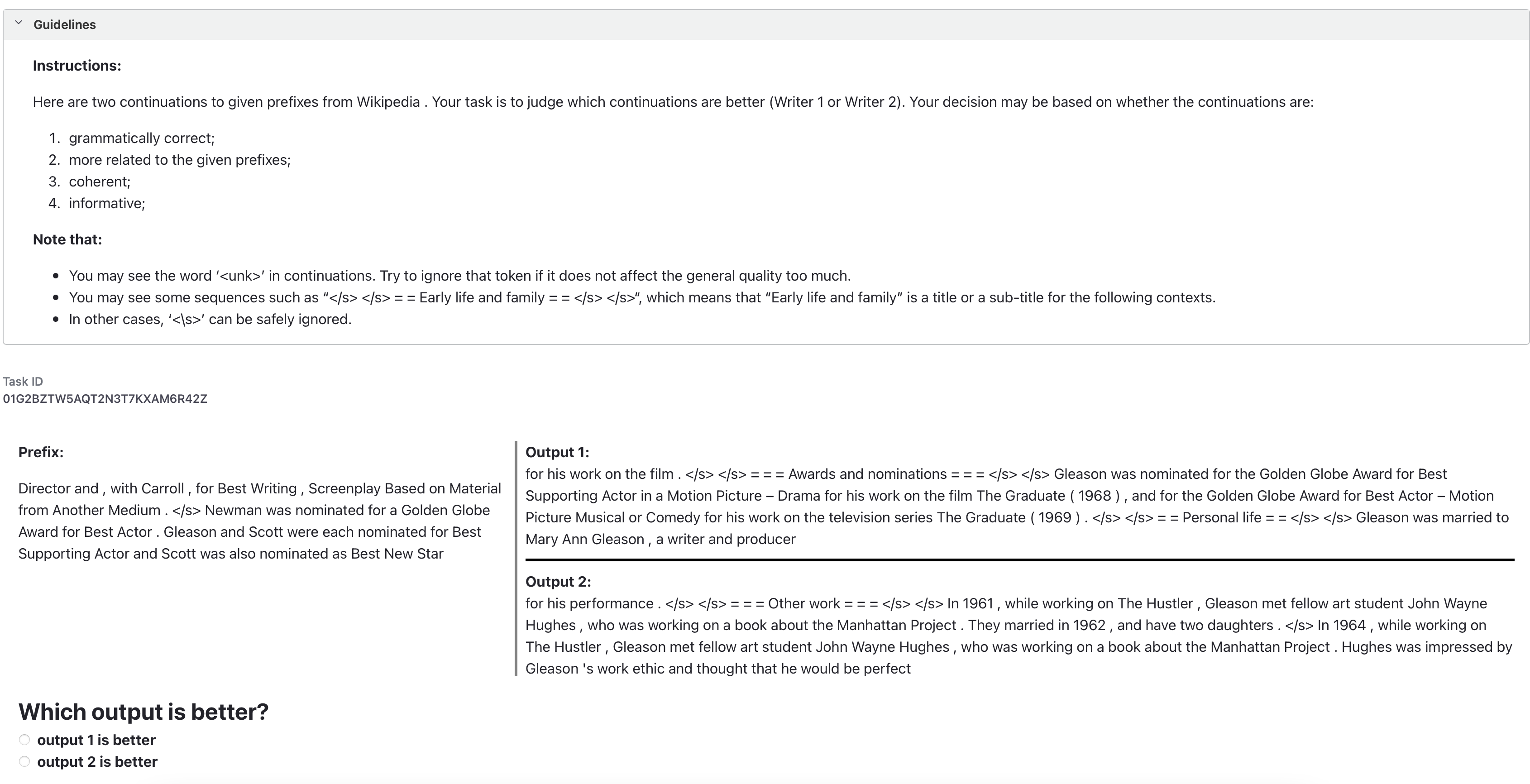} \\
\caption{Screenshot of the user interface in human evaluation experiments.}
\label{fig:human_eval_interface}
\end{figure}

We conduct a pairwise crowdworker evaluation to judge the quality of the generations of DITTO compared to other baselines. Models generate continuations based on the same 100 random prefixes from the test set of Wikitext-103. For each comparison of two continuations, three fluent English speakers are asked to independently judge which continuation is better. The overall quality is judged from four aspects: 1) grammar correctness; 2) relevance; 3) coherence; and 4) informativeness. The win rate is calculated as the total number of times that DITTO beats the other model divided by the total number of comparisons in the evaluation. The interface for the human evaluation is shown in Figure~\ref{fig:human_eval_interface}.

\section{Examples}~\label{appendix:examples}
In this section, we show the examples of generated texts of different training methods in the open-ended generation task. The examples are presented in the following tables.

\begin{table}[htb]
\centering
\caption{Example 1 of generated texts by models with greedy decoding.}
\label{table:example1}

\begin{tabularx}{\textwidth}{X}
\toprule
\textbf{Prefix}  \\\midrule
was an immediate public outcry following the publication of the reports , particularly with regards to the decision to not repatriate the bodies of the dead . The reports generated considerable discussion in the press which ultimately led to a heated debate in Parliament on 4 May 1920 . Sir
\\\midrule
\textbf{MLE}\\\midrule
Edward Cave , the Director of the Public Prosecutions , was quoted as saying that Cave had " no idea what the truth was , and that the truth was not the truth. But what the truth was , and that the truth was not the truth . But what the truth was , and that the truth was not the truth . But what the truth was , and that the truth was not the truth . </s> But what the truth was , and that the truth was not the truth . But what the
\\\midrule
\textbf{UL-token}
\\\midrule
John Russell , who had been appointed High Commissioner for Ireland , was also present at the hearing . </s> </s> = = = = Public response = = = = </s> </s> The public response to the report was mixed . The Irish Times called it " a most damning and comprehensive report on the Irish Question " . The Irish Times called it " a book of great importance " and " a work of great importance " . The Irish Times called it " a work of great importance " and " a work of great importance
\\\midrule
\textbf{UL-token+seq}\\\midrule
Geoffrey Howe , the Prime Minister , and his wife were among the signatories to the London Necropolis Act , which was passed on 30 June 1921 . </s> </s> = = = = The London Necropolis Railway Act 1921 = 6 1 ⁄ 2-mile ( 8.0 km ) long railway from Brookwood to Brookwood was the first railway in the world to be built in the United States outside the British Empire . The Act allowed the LSWR to carry out the work by rail from its own station at Brookwood , and the LSWR
\\\midrule
\textbf{SG}\\\midrule
Edward Cave read the report and stated that he had been " deeply disappointed " that the government had not acted on the grounds that it had not acted on the grounds that it had not acted on the grounds that it had not acted on the grounds that it had not acted on the grounds that it had not acted on the grounds that it had not acted on the grounds that it had not acted on the grounds that it had not acted on the grounds that it had not acted on the grounds that it had not
\\\midrule
\textbf{DITTO}\\\midrule
Edward Cave , the Director of Public Prosecutions , was quoted as saying that " the Government has not been able to make a decision on the issue of the bodies of the dead . " </s> </s> = = = Public reaction = = = </s> </s> The public reaction to the killings was generally positive . The Times called the actions " a most appalling and appalling act " and the Daily Mail called for the police to be withdrawn . The Daily Mail called for the police to be withdrawn , but the Daily Mail said that
\\
\bottomrule
\end{tabularx}

\end{table}

\begin{table}[htb]
\centering
\caption{Example 2 of generated texts by models with greedy decoding.}
\label{table:example2}

\begin{tabularx}{\textwidth}{X}
\toprule
\textbf{Prefix}  \\\midrule
is summed up as " not to be missed . " </s> Paul Newman reprised his role as Fast Eddie Felson in the 1986 film The Color of Money , for which he won the Academy Award for Best Actor in a Leading Role . A number of observers and
\\\midrule
\textbf{MLE}\\\midrule
critics have praised Fast Eddie Felson as one of the greatest actors of all time . In 2007 , Fast Eddie Felson was ranked \# 1 on Empire magazine 's list of the 500 Greatest Movie Characters of All Time . In 2010 , Fast Eddie 's popularity was listed at \# 3 on Empire 's list of the 500 Greatest Movie Characters of All Time . In 2012 , Fast Eddie 's popularity was ranked \# 1 on Empire 's list of the 500 Greatest Movie Characters of All Time . In 2013 , Fast Eddie 's popularity was
\\\midrule
\textbf{UL-token}
\\\midrule
critics praised his performance . The New York Times ' Janet Maslin wrote , " Fast Eddie Felson is a real joy to watch , and he is a real joy to watch . " The New York Times ' Janet Maslin wrote , " Fast Eddie Felson is a real joy to watch , and he is a real joy to watch . " The New York Times ' Janet Maslin wrote , " Fast Eddie Felson is a real joy to watch , and he is a real joy to watch . " The New York Times
\\\midrule
\textbf{UL-token+seq}\\\midrule
critics have praised Fast Eddie Felson as the film 's most memorable moment . In his review for the Chicago Reader , critic Richard Schickel wrote that " Fast Eddie Felson is the first of the great actors to come across as a master of the art of acting . " In his review for the New York Times , critic Anthony Boucher wrote that " Fast Eddie Felson is the first actor who can make a living through the motions of his actors , and the film is a triumph of invention . " In his review for the
\\\midrule
\textbf{SG}\\\midrule
critics have praised Fast Eddie 's performance , including Roger Ebert , who wrote that " Fast Eddie is a great comic actor , and he has a great comic timing . " </s> </s> = = Personal life = = </s> </s> Fast Eddie was married twice , first to Patricia ( née <unk> ) and then to Patricia ( née <unk> ) . They divorced in 1977 . Their son , Eddie Jr . , is a retired basketball player who played for the Denver Nuggets , New York Knicks , Boston Celtics , and Dallas Mavericks .
\\\midrule
\textbf{DITTO}\\\midrule
critics have praised Fast Eddie 's performance , including the New York Times critic Bosley Crowther , who wrote that " Fast Eddie is a fine actor , and he is a fine actor . " </s> </s> = = = Accolades = = = </s> </s> Fast Eddie was nominated for the Academy Award for Best Actor in a Leading Role for his role as the title character in the 1986 film Fast Eddie . He was also nominated for the Golden Globe Award for Best Actor – Motion Picture Musical or Comedy for his role as the title
\\
\bottomrule
\end{tabularx}

\end{table}

\begin{table}[htb]
\centering
\caption{Example 1 of generated texts by models with nucleus sampling.}
\label{table:example4}

\begin{tabularx}{\textwidth}{X}
\toprule
\textbf{Prefix}  \\\midrule
's screenplay was selected by the Writers Guild of America in 2006 as the 96th best motion picture screenplay of all time . In June 2008 , AFI released its " Ten top Ten " — the best ten films in ten " classic " American film genres — after
\\\midrule
\textbf{MLE}\\\midrule
a deal with Chase \& Caldera Entertainment was reached . Ten countries also released 100 top ten films in multiple-day international markets . </s> </s> = = = 2010s = = = </s> </s> </s> = = = = 2010s = = = = </s> </s> </s> = = = = = 2010s = = = = = </s> </s> </s> = = = = = 2010s = = = = = </s> </s> </s> = = = = = = 3D movies = = = = = = </s> </s> </s> = = = = 3D movies
\\\midrule
\textbf{UL-token}
\\\midrule
adjusting their publication guidelines for rated films and having it between 25th and 26th , with cumulative scores of 70.2 and 90.5 . In December 2011 , The Hollywood Reporter said that with those achieved in reverse , characters such as Hopper could become " more powerful than previously unseen " . It also said that women would have to look after the characters in pictures like Barbie and Paris Hilton for inspiration . </s> In April 2014 , Warner Bros. was sued by Roger Hathaway , David Fincher , Arnold Schwarzenegger , Darren Aronofsky ,
\\\midrule
\textbf{UL-token+seq}\\\midrule
polling for AFI 's 10 Top 10 , and also including WALL-E , WALL Scattered , WALL Building 3 , and WALL Jack . </s> In October 2013 , Netflix announced the long-running website Expedition to Antarctica , celebrating the new Space Travel , and the search for and rescue mission planners had been in the middle of the year . A 2012 study by video game journalist Toby Philpott found a large majority of the space explorers who have survived on the ship as well as exploring and exploring locations . The Discovery and Curiosity
\\\midrule
\textbf{SG}\\\midrule
polling over 300 people from 108 countries worldwide . In 2012 , AFI ranked 65th on Bravo 's 100 Years ... 100 Movies list . </s> </s> = = Awards = = </s> </s> Writers Guild of America 's 50 Years ... 100 Heroes \& Villains : </s> \^ * AFI 's 100 Years ... 100 Movie Quotes : " Run to your sister , son ... If you hate her [ ... ] you wouldn 't believe that you 're too stupid ? " – Nominated </s> AFI 's 100 Years ... 100 Thrills – Nominated </s> AFI 's
\\\midrule
\textbf{DITTO}\\\midrule
polling more than 130,000 people at the screening and success . </s> In October 2009 , AFI 's 10 Top 10 ranked " The 100 Scariest Movie Trailer " , the first-ever list of AFI 's 100 Years ... 100 Thrills — while the film was also ranked among the top ten all-time greatest movie villains in the last five years — and in December 2012 , Time magazine listed it as the sixth-greatest film villain in history . In August 2012 , Total Film ranked " The 100 Scariest Movie Characters
\\
\bottomrule
\end{tabularx}

\end{table}

\begin{table}[htb]
\centering
\caption{Example 2 of generated texts by models with nucleus sampling.}
\label{table:example5}

\begin{tabularx}{\textwidth}{X}
\toprule
\textbf{Prefix}  \\\midrule
business activities upon his return . He invested in a wide variety of businesses and infrastructure , spurring economic activity in the state . His investments ranged widely , including maritime insurance ( where is father-in-law had made his fortune ) , bridges , locks ,
\\\midrule
\textbf{MLE}\\\midrule
power plants and homes . He and his wife died from injuries sustained in labor ventures when they were old . He was buried in Père Lachaise Cemetery in 1919 . He left a lasting legacy to his family . </s> After leaving politics , Holloway became close friends with Josiah Quincy III . </s> </s> = = Early life and family = = </s> </s> Holloway was born on 12 November 1827 in Saint Joseph , Missouri , the only child of Elias and Emily Holloway . Most of Holloway 's family was derived from him , though they
\\\midrule
\textbf{UL-token}
\\\midrule
dams , gambling equipment , and positions . </s> Henderson lost heavily during his sojourn to Mississippi . He attempted to build a canal , which he provided on several occasions with investment in agriculture and other pursuits . His wife Grace later wrote that Henderson " loathed working men so long that it nearly destroyed the whole business " . She only learned to gamble at all levels of the United States in the late 1840s . When he had returned to Columbus in the mid-1830s , all those who had survived the Ohio River had little
\\\midrule
\textbf{UL-token+seq}\\\midrule
a gas leak , and the offices of the administration , engineers , and messengers for scientific and industrial interests . He also invested in William <unk> Ingram who purchased two berths from the Sons of the American Revolution , a natural gas ship that had been fired in 1807 in retaliation for the order to resupply the British Army troops at Fredericksburg . He did not work on the reorganization of the army , but supported the need for self-government-managers in a transition known as Federal service . </s> Small farmers raised cattle in their
\\\midrule
\textbf{SG}\\\midrule
tunnels , pumps , fire stations , telephones , showers and ships . </s> Retaining custody after the September 11 attacks , Romney returned to family life by appointing first-class city councilman Carl McCall as U.S. Attorney General . Later that year , he was recognized by the Sons of Prosperity organization , where he served as chair of the Drafting Committee on Government Reform . While in the mid-1980s , Romney unsuccessfully ran for re-election as Secretary of Commerce under Governor Lawrence B. Jennings , losing to Richards . While in office ,
\\\midrule
\textbf{DITTO}\\\midrule
hotels , art galleries , and utilities . He formed new , larger enterprises ( offering bonds and furniture ) , offering services to smaller companies , attracted much investment in newly discovered companies . He also served as the state 's commissioner of the Board of Higher Education , providing education and higher education . </s> </s> = = Personal life = = </s> </s> John Smith was born in 1830 in Salt Lake City . His father , also named John , was an army officer stationed in a successful military career in the early 19th century .
\\
\bottomrule
\end{tabularx}

\end{table}

\end{document}